\documentclass{article} %
\usepackage{iclr2027_conference,times}

\newif\ifpreprint
\preprinttrue
\ifpreprint
\iclrfinalcopy
\fi

\usepackage[utf8]{inputenc} %
\usepackage[T1]{fontenc}    %
\usepackage{hyperref}       %
\usepackage{url}            %
\usepackage{booktabs}       %
\usepackage{amsfonts}       %
\usepackage{nicefrac}       %
\usepackage{microtype}      %
\usepackage{xcolor}         %
\usepackage{graphicx}
\usepackage{wrapfig}
\usepackage{multirow}
\usepackage{amsmath}

\newif\ifshowcomments
\showcommentstrue  %

\ifshowcomments

    \newcommand{\caroline}[1]{{\textcolor{blue}{[caroline: #1]}}}

\else
    \newcommand{\caroline}[1]{}
\fi

\newif\ifshowrevisions
\showrevisionstrue
\ifshowrevisions
    
\else
    
\fi

\expandafter\ifx\csname ifpreprint\endcsname\relax
  \expandafter\newif\csname ifpreprint\endcsname
\fi

\usepackage{placeins} %
\usepackage{float}
\usepackage{makecell}
\usepackage{wrapfig}
\usepackage{graphicx}
\usepackage{subcaption}
\usepackage{enumitem}
\usepackage{enumitem}
\usepackage{wrapfig}
\usepackage{siunitx}
\usepackage{setspace}
\usepackage{cleveref}
\usepackage{url}

\crefname{algorithm}{Algorithm}{Algorithms}
\crefname{table}{Table}{Tables}
\crefname{line}{Line}{Lines}

\definecolor{asparagus}{rgb}{0.53, 0.66, 0.42}
\definecolor{bittersweet}{rgb}{1.0, 0.44, 0.37}
\definecolor{ao(english)}{rgb}{0.0, 0.5, 0.0}
\definecolor{mydarkblue}{rgb}{0,0.08,0.45}
\definecolor{blue(ncs)}{rgb}{0.0, 0.53, 0.74}
\definecolor{celestialblue}{rgb}{0.29, 0.59, 0.82}
\definecolor{earthyellow}{rgb}{0.88, 0.66, 0.37}
\definecolor{lightgray}{rgb}{0.83, 0.83, 0.83}
\definecolor{brickred}{rgb}{0.8, 0.25, 0.33}
\definecolor{myblue}{rgb}{0.345, 0.545, 0.902}
\ifdefined\nohyperref\else\ifdefined\hypersetup
  \hypersetup{ %
    pdftitle={},
    pdfsubject={},
    pdfkeywords={},
    pdfborder=0 0 0,
    pdfpagemode=UseNone,
    colorlinks=true,
    linkcolor=mydarkblue,
    citecolor=ao(english),
    filecolor=mydarkblue,
    urlcolor=magenta,
    }
  \fi
\fi

\usepackage{amsmath,amsfonts,bm}

\def\eqref#1{equation~\ref{#1}}

\def\1{\bm{1}}

\DeclareMathAlphabet{\mathsfit}{\encodingdefault}{\sfdefault}{m}{sl}
\SetMathAlphabet{\mathsfit}{bold}{\encodingdefault}{\sfdefault}{bx}{n}


\title{JaxAHT: A JAX-Based Library for Ad Hoc Teamwork}

\author{
    Caroline Wang\textsuperscript{1}\thanks{\quad Corresponding author: \texttt{caroline.l.wang@utexas.edu}} \quad Rolando Fernandez\textsuperscript{1} \quad
    Zelal Su Mustafaoglu\textsuperscript{1} \quad Montek Kundan\textsuperscript{1} \\
    \textbf{Johnny Liu}\textsuperscript{1} \quad 
    \textbf{Aditya Madhan}\textsuperscript{1} \quad
    \textbf{Zhihan Wang}\textsuperscript{1} \quad 
    \textbf{Lingyun Xiao}\textsuperscript{1} \quad
    \textbf{Di Yang Shi}\textsuperscript{1} \\
    \textbf{Jiaxun Cui}\textsuperscript{1,2} \quad 
    \textbf{Arrasy Rahman}\textsuperscript{1} \quad 
    \textbf{Peter Stone}\textsuperscript{1,3} \\
    \textsuperscript{1}Department of Computer Science, The University of Texas at Austin \\
    \textsuperscript{2}Google
}

\begin{document}

\maketitle
\ifpreprint\lhead{Preprint}\fi

\begin{abstract}
As autonomous agents are increasingly deployed, they must be able to coordinate with previously unseen agents, such as humans. This challenge has been formalized as Ad Hoc Teamwork (AHT), and has  attracted a substantial body of research. However, progress has been hindered by the prohibitive computational cost of the AHT research lifecycle, lack of standardized benchmark implementations, and the absence of a diverse, validated evaluation teammate suite. This work introduces \textbf{JaxAHT}, the first open-source, JAX-based library designed to accelerate and standardize the AHT research lifecycle. Leveraging JAX's hardware acceleration and massive parallelization capabilities, JaxAHT provides a unified framework for teammate generation, ego agent training, and evaluation against unseen teammates, achieving approximately 95x wall-clock speedup over existing PyTorch counterparts. Along with the library, we contribute a diverse suite of evaluation teammates across Level-Based Foraging, Overcooked, and Hanabi. Using JaxAHT, we conduct a large-scale, compute-controlled benchmark study of AHT algorithms, finding that no algorithm consistently performs best, that agent modeling primarily offers benefits in role-based scenarios, and that best response returns are underestimated by reinforcement learning from scratch.
\end{abstract}

\vspace{-.3cm}
\section{Introduction}
\label{sec:introduction}
\vspace{-.3cm}
Autonomous agents in the real world must possess the social intelligence to work with other agents in shared environments, with whom they may not share conventions or communication protocols~\citep{stone2010adhoc, barrett2017makingfriends}. 
For example, robots from different manufacturers may be deployed together in a warehouse setting, or an assistant agent may be paired with humans it has never met~\citep{mirsky2022survey}. 

Ad Hoc Teamwork (AHT) formalizes this challenge by requiring an \textit{ego agent} to coordinate effectively with teammates whose behaviors are competent yet unknown~\citep{stone2010adhoc}. 
In contrast to cooperative multi-agent reinforcement learning (MARL), where teams are trained together from scratch and often converge on arbitrary conventions that break down with new partners~\citep{carroll_utility_2019,hu2020otherplay}, AHT agents must adapt on the fly to whoever they are paired with.
The design and evaluation of AHT agents is particularly challenging, as both training and evaluation depend on constructing diverse, representative sets of teammates. 

We identify three critical bottlenecks of AHT research. 
First, computational cost severely limits research iteration speed and rigor. 
The \textit{AHT research lifecycle} comprises generating diverse teammate populations, training ego agents to cooperate with these teammates, and evaluating their generalization to held-out teammates~\citep{rahman2023brdiv,papoudakis2020liam}. 
These stages are computationally prohibitive under existing community PyTorch frameworks~\citep{paszke2019pytorch}, limiting iteration speed and adherence to ideal experimental practices such as hyperparameter searches. 
Second, lack of standard evaluation teammates leads researchers to design their own with limited diversity validation, leaving evaluation thoroughness in question and results non-comparable across publications.
Third, the lack of standardized implementations combined with the complexity of the AHT research lifecycle forces researchers to orchestrate disparate algorithm implementations, raising the barrier to entry and limiting investigation of methods that combine components of the AHT problem. \looseness=-1 %

This work introduces \textbf{JaxAHT}, a JAX-based library that mitigates all three bottlenecks. By leveraging JAX's hardware acceleration and massive parallelization  capabilities~\citep{jax2018github}, 
JaxAHT provides a unified, accelerated AHT library and a dataset of diverse evaluation teammates across standard benchmark tasks. The contributions  of the paper are:

\begin{itemize}[noitemsep, leftmargin=*]
    \item \textbf{Unified JAX-Based AHT Library:} JaxAHT is the first library to support the full AHT research lifecycle, providing standardized JAX implementations of teammate generation, ego agent training, and cooperative MARL algorithms within one unified framework. JaxAHT achieves  approximately 95x wall-clock time speedup over community PyTorch implementations.
    \item \textbf{Diverse Evaluation Teammate Suite:} 
    We provide a diverse suite of heuristic, RL-based, and human-data-derived evaluation teammates for the standard AHT benchmark tasks of Level-Based Foraging (LBF), Overcooked, and Hanabi, and validate its quality and diversity via empirical cross-play returns and behavioral embeddings (Sec.~\ref{sec:exp:teammates}). We also present an open-source human gameplay dataset for LBF (1299 games from 180 participants; App.~\ref{app:human_data_lbf}).
    \item \textbf{Benchmark Study:} We benchmark all algorithms on the JaxAHT evaluation teammates under controlled compute budgets with per-algorithm hyperparameter sweeps, demonstrating the large-scale empirical analysis JaxAHT enables and providing insights on the state of AHT research.
\end{itemize}

Together, these contributions enable large-scale AHT studies that were previously intractable.
Code and datasets of pre-generated teammates and human gameplay are open-sourced (App.~\ref{app:code_data}). \looseness=-1 %

\vspace{-0.1cm}
\section{Related Work}
\label{sec:related_work}
\vspace{-0.1cm}

\begin{table}[t]
    \centering
    \footnotesize
    \setlength{\tabcolsep}{2.5pt}
    \renewcommand{\arraystretch}{1.0}
    \begin{tabular}{lcccccc}
        \toprule
        Library & JAX & Multi-Env & Ego Algs. & Team Gen. & Eval Teammates & Human Data \\
        \midrule
        JaxAHT
        & \checkmark & \checkmark & \checkmark & \checkmark & \checkmark & \checkmark \\
        ZSCEval~\citep{wang2024zsceval}
        &            & \checkmark &            & \checkmark & \checkmark & \checkmark \\
        AH2AC2~\citep{dizdarevic2025ah2ac2}
        & \checkmark &            &            &            &            & \checkmark \\
        ICRL4AHT~\citep{jing2026icrl4aht}
        & \checkmark &            & \checkmark & \checkmark & \checkmark &            \\
        Assistax~\citep{hinckeldey2026assistax}
        & \checkmark & \checkmark &            &            & \checkmark &            \\
        \bottomrule
    \end{tabular}
    \caption{\small\textbf{Library comparison.}
    JaxAHT is the only library that covers the full AHT research lifecycle.}
    \label{tab:library_comparison}
\end{table}

We provide an overview of related AHT evaluation frameworks and JAX-based RL libraries.
\vspace{-.3cm}
\paragraph{Benchmarks and Evaluation for AHT.}
Environments such as Overcooked~\citep{carroll_utility_2019,gessler2025overcookedv2} and Hanabi~\citep{bard2020hanabi} are widely used to evaluate AHT methods, yet few libraries support multiple AHT algorithms and environments jointly~\citep{mirsky2022survey}. 
The most closely related works are ZSCEval~\citep{wang2024zsceval}, AH2AC2~\citep{dizdarevic2025ah2ac2}, ICRL4AHT~\citep{jing2026icrl4aht}, and Assistax~\citep{hinckeldey2026assistax} (Table~\ref{tab:library_comparison}).
ZSCEval focuses on teammate generation algorithms and proposes BRDiv for partner selection and BRProx for measuring generalization, but is implemented in PyTorch, which limits the scale of experiments feasible at academic compute budgets, and does not provide ego agent training algorithms. 
AH2AC2 focuses exclusively on Hanabi, providing gold-standard human proxy agents trained on approximately 100k human games, but gates access behind an evaluation API.
ICRL4AHT is a JAX-based benchmark of in-context adaptation to unseen teammates and layouts in Overcooked-v2~\citep{gessler2025overcookedv2}, providing teammate generation algorithms and a standardized teammate suite, but for a single environment and with ego agent baselines limited to in-context RL methods.
Assistax provides JAX-accelerated assistive robotics tasks with an evaluation population of humanoid partners, but does not include AHT algorithms.
More broadly, the Overcooked Generalisation Challenge~\citep{ruhdorfer2025ogc} addresses cooperation with unseen partners in procedurally generated Overcooked layouts only, while JaxAHT addresses multiple types of tasks. 
SocialJax~\citep{guo2026socialjax} targets sequential social dilemmas, rather than the AHT problem.
JaxAHT complements these benchmarks by providing shared interfaces for the complete AHT lifecycle across multiple environments, enabling comparisons across algorithm classes via a standardized teammate suite, and performing a benchmark study.
\vspace{-.3cm}
\paragraph{JAX-based RL Libraries.}
JAX~\citep{jax2018github} enables GPU-accelerated numerical computing through JIT compilation and auto-vectorization. 
PureJaxRL~\citep{lu2022discovered} demonstrated that end-to-end JAX implementations, in which environments, policies, and training loops all execute on the GPU, achieve over $1000\times$ speedups over PyTorch-based approaches when parallelizing over many environments and seeds, using minimal single-file implementations of core RL algorithms.
For MARL research, JaxMARL~\citep{rutherford2024jaxmarl} provides JAX-based environments and algorithms, while Jumanji~\citep{bonnet2023jumanji} provides JAX-based environments featuring industry-oriented decision problems. 
NiceWebRL~\citep{carvalho2025nicewebrl} extends JAX environments to support web-based human subject experiments.
\vspace{-.3cm}

\section{Problem Setting and Design Philosophy}
\label{sec:bg}
\vspace{-0.2cm}

AHT studies the fully cooperative setting in which an ego agent must coordinate with teammates whose behaviors are unknown in advance and with whom it cannot pre-coordinate \citep{stone2010adhoc, mirsky2022survey}. 
We model the AHT problem as an
$N$-agent decentralized partially observable Markov decision process~\citep[Dec-POMDP;][]{Bernstein2002DecPOMDP}, $\mathcal{M} = \langle \mathcal{S}, \{\mathcal{A}_i\}_{i=1}^{N}, P, \{\mathcal{O}_i\}_{i=1}^{N}, \Omega, r, \rho, \gamma \rangle$.
Each agent $i$ selects an action $a_i^t$ conditioned on its local action-observation
history $\tau_i^t=(o_i^0,a_i^0,\dots,o_i^t)$ according to a decentralized policy $\pi_i(a_i^t\mid\tau_i^t)$, and receives a shared team reward. 
The joint policy is denoted by $\boldsymbol{\pi}=(\pi_1,\dots,\pi_N)$. \looseness=-1 %

Let $e \in \{1,\dots,N\}$ denote the ego agent and let
$\boldsymbol{\pi}_{-e}$ denote the joint policy of the $N-1$ teammates the ego agent is paired with, which are drawn from $\mathcal{T}_{\mathrm{train}}$ during training and from $\mathcal{T}_{\mathrm{eval}}$ at evaluation time. Following standard AHT
evaluation practice, the ego agent is trained to cooperate broadly and then evaluated with previously
unseen teammates drawn from a held-out set $\mathcal{T}_{\mathrm{eval}}$
\citep{papoudakis2020liam,rahman2023brdiv,wang2024nagent}:
\begin{equation}
\label{eq:aht_objective}
J_{\mathrm{AHT}}(\pi_e)
=
\mathbb{E}_{\boldsymbol{\pi}_{-e}\sim\mathcal{P}_{\mathrm{eval}}}\left[J(\pi_e,\boldsymbol{\pi}_{-e})\right],
\end{equation}
where $J(\pi_e,\boldsymbol{\pi}_{-e})$ is the expected discounted team return, and $\mathcal{P}_{\mathrm{eval}}$ is a distribution over $\mathcal{T}_{\mathrm{eval}}$, taken as uniform per common practice, so that Eq.~(\ref{eq:aht_objective}) reduces to the mean return over evaluation teammates. \looseness=-1 %

A useful reference point for a fixed teammate is its \emph{best response} (BR), $\mathrm{BR}(\boldsymbol{\pi}_{-e}) \in \arg\max_{\pi} J(\pi,\boldsymbol{\pi}_{-e})$,  the policy that achieves the highest return with teammates $\boldsymbol{\pi}_{-e}$.
Exact BRs are generally intractable in large Dec-POMDPs, but may be approximated with an ego agent training or MARL algorithm.
BR reasoning is central to prior AHT work for constructing teammate populations~\citep{albrecht_game-theoretic_2013,strouse2021fcp,rahman2023brdiv,rahman2024lbrdiv}.
As described in Sec.~\ref{sec:eval_metrics}, JaxAHT uses BR returns to normalize evaluation returns across teammates.

\begin{figure}[t]
    \centering
    \includegraphics[width=0.8\textwidth]{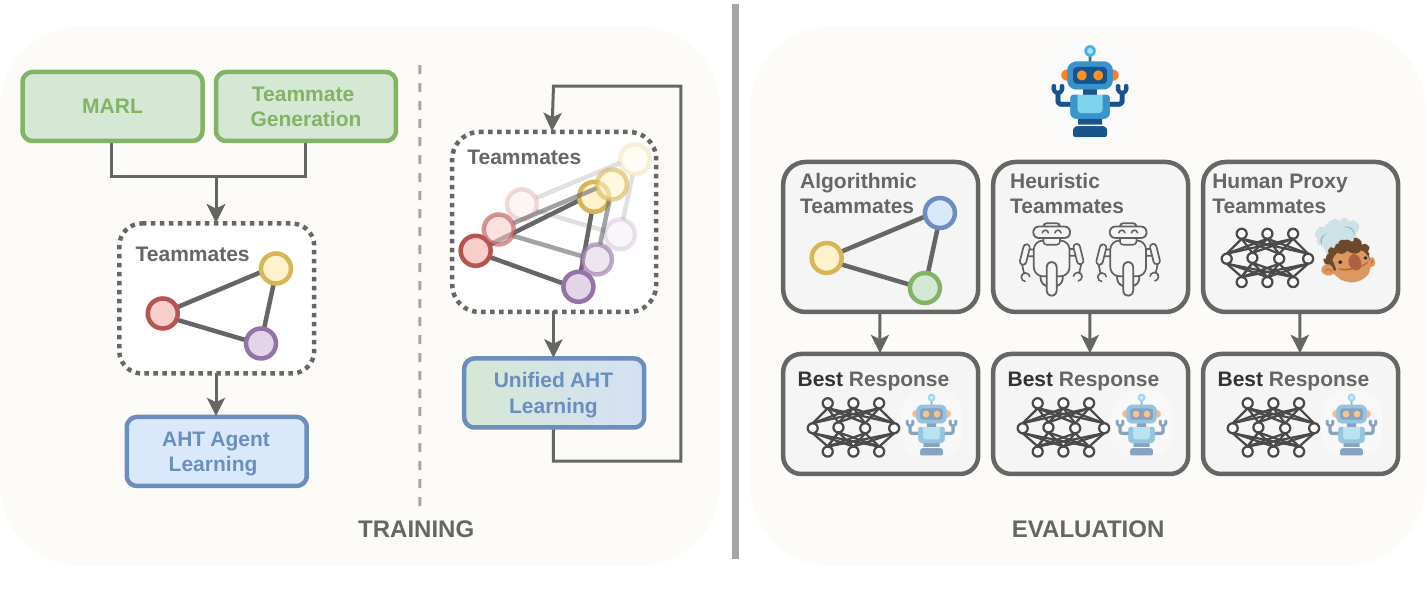}
    \caption{\textbf{JaxAHT Design Philosophy.} JaxAHT unifies the full AHT research lifecycle within a shared codebase. During training, common agent and population abstractions support MARL, teammate-generation, ego-agent learning, and unified AHT methods that jointly construct training teammates and ego policies. During evaluation, learned agents are tested against a standardized suite of heuristic, algorithmic, and human-data-informed teammates, together with BR references, enabling consistent and scalable benchmarking across AHT methods.}
    \label{fig:design_philosophy}
\end{figure}

Addressing the AHT problem involves three stages, each corresponding to a component of JaxAHT.
First, a set of training teammates $\mathcal{T}_{\mathrm{train}}$ must be constructed, either by hand or by relying on \textit{teammate generation} or \textit{cooperative MARL} algorithms.
Second, the ego policy $\pi_e$ is learned by interacting with teammates sampled from $\mathcal{T}_{\mathrm{train}}$, using an \textit{ego agent training} or cooperative MARL algorithm.
Third, the resulting ego policy must be evaluated against a standardized \textit{evaluation set} of teammates.
These stages of AHT research rely on three main algorithm types: teammate generation, ego agent training, and cooperative MARL.
The first two are AHT algorithm types~\citep{mirsky2022survey}, while the third has conventionally been studied as its own problem~\citep{marl-book-albrecht24,yuan2023survey}, although all three may support various stages of AHT research.
Research on one type often requires the others; for instance, evaluating a teammate generation method requires an ego agent training method. \looseness=-1 %

JaxAHT provides a unified interface for all three, enabling research on individual components or their combinations, as illustrated in Fig.~\ref{fig:design_philosophy}.
These components also support AHT algorithms \textit{beyond} the conventional two-stage framework, such as \textit{unified} methods that jointly generate teammates and ego agents, built from the same agent and population abstractions and MARL procedures for training BRs. \looseness=-1 %

For fast iteration and clean code, we follow the single-file model of JaxMARL and CleanRL~\citep{huang2022cleanrl}, but \textit{minimally modularize}: algorithms are implemented in single files that are easy to understand and extend, while a shared agent and population interface allows agents trained by one algorithm to be used by another---a common AHT research workflow.
The library also supports \texttt{hydra} configuration management~\citep{yadan2019hydra} and WandB logging~\citep{wandb}, facilitating reproducibility and research collaboration.

\vspace{-0.2cm}
\section{JaxAHT Library}
\label{sec:jaxaht}
\vspace{-0.2cm}

JaxAHT addresses the AHT research bottlenecks described in Sec.~\ref{sec:introduction} with JAX-accelerated algorithms and environments supporting all three lifecycle stages (Sec.~\ref{sec:bg}), and a standardized evaluation set of teammates.
This section describes the algorithms, environments,  evaluation teammates, and metrics.

\paragraph{Algorithms}
\setlength{\columnsep}{6pt}
\begin{wraptable}[14]{r}{0.5\textwidth}
    \centering
    \vspace{-6pt}
    \setlength{\tabcolsep}{4pt}
    \resizebox{\linewidth}{!}{%
        \begin{tabular}{ll}
            \toprule
            \textbf{Category} & \textbf{Algorithm} \\
            \midrule
            \textbf{Ego Agent Training}
                & PPO~\citep{Schulman2017ProximalPO} \\
                & LIAM~\citep{papoudakis2020liam} \\
                & MeLIBA~\citep{zintgraf2021meliba} \\
            \textbf{Teammate Generation}
                & FCP~\citep{strouse2021fcp} \\
                & BRDiv~\citep{rahman2023brdiv} \\
                & LBRDiv~\citep{rahman2024lbrdiv} \\
                & CoMeDi~\citep{sarkar2023comedi} \\
            \textbf{Unified AHT Training}
                & TrajeDi~\citep{pmlr-v139-lupu21a} \\
                & COLE~\citep{li2023cole} \\
                & ROTATE~\citep{wang2025rotateregretdrivenopenendedtraining} \\
            \midrule
            \textbf{General MARL}
                & IPPO~\citep{dewitt2020independentlearning} \\
            \bottomrule
        \end{tabular}%
    }
    \vspace{3pt}
    \caption{\small\textbf{JaxAHT Algorithms.}}
    \label{tab:algo_categories}
\end{wraptable}

Table~\ref{tab:algo_categories} summarizes the algorithms currently implemented in JaxAHT.
Each is briefly described below and in more detail in App.~\ref{app:alg_overview}.
We also release tuned hyperparameters for each algorithm on the benchmarked tasks (Sec.~\ref{sec:exp:benchmark}), providing performant baseline configurations for future research.

For \textit{teammate generation}, we implement Fictitious Co-Play~\citep[FCP;][]{strouse2021fcp}, which generates diverse teammates by varying the seed and storing multiple checkpoints, and population-based algorithms that maximize diversity through adversarial objectives, such as BRDiv~\citep{rahman2023brdiv}, its extension LBRDiv~\citep{rahman2024lbrdiv}, and CoMeDi~\citep{sarkar2023comedi}.
For \textit{ego agent training}, we support PPO~\citep{Schulman2017ProximalPO} and the agent modeling methods LIAM~\citep{papoudakis2020liam} and MeLIBA~\citep{zintgraf2021meliba}, 
which extend PPO by conditioning the ego policy on latent representations of the teammate inferred from interaction history.
For \textit{MARL}, we include IPPO~\citep{dewitt2020independentlearning} as the sole exemplar method.
We refer researchers seeking a broader range of MARL algorithms to JaxMARL~\citep{rutherford2024jaxmarl}, which JaxAHT builds upon.
Finally, we support \textit{unified training} methods that jointly train ego agents and teammate populations. 
These include TrajeDi~\citep{pmlr-v139-lupu21a}, which maximizes population diversity via trajectory-based information-theoretic objectives; COLE~\citep{li2023cole}, which uses graph-theoretic notions of cooperative compatibility to define a sampling distribution over the population and iteratively trains a BR to it; and ROTATE~\citep{wang2025rotateregretdrivenopenendedtraining}, an open-ended method that alternates between generating competent teammates that maximize the ego agent's cooperative regret and training the ego agent against the accumulated population.

\paragraph{Environments}
Table~\ref{tab:env_overview} summarizes supported environments.
We build on the growing ecosystem of JAX-based implementations of commonly used AHT environments~\citep{bonnet2023jumanji,rutherford2024jaxmarl}.
We select the following environments for their prevalence in AHT research and their support for diverse cooperative conventions: LBF~\citep{albrecht_game-theoretic_2013, bonnet2023jumanji}, Overcooked-v1~\citep{carroll_utility_2019}, and Hanabi~\citep{bard2020hanabi,rutherford2024jaxmarl}.

\begin{table}[t]
    \centering
    \begin{tabular}{lll p{4cm}}
        \hline
        Environment & Source & Eval Teammates & Variants \\
        \hline
        \textbf{Level-Based Foraging} &
        \href{https://github.com/instadeepai/jumanji}{Jumanji} &
        $\checkmark$ &
        7x7 grid, constant levels\\
        &&& 12x12 grid \\
        \textbf{Overcooked-v1} &
        \href{https://github.com/FLAIROx/JaxMARL}{JaxMARL} &
        $\checkmark$ &
        Asymmetric Advantages\\
        &&& Coordination Ring\\
        &&& Counter Circuit\\
        &&& Cramped Room\\
        &&& Forced Coordination \\
        \textbf{Hanabi} &
        \href{https://github.com/FLAIROx/JaxMARL}{JaxMARL} &
        $\checkmark$ &
        Full (5 colors $\times$ 5 ranks) \\
        &&& Mini (3 colors $\times$ 3 ranks) \\
        \hline
    \end{tabular}
    \caption{\small\textbf{JaxAHT Environments.} Overview of environments supported by JaxAHT.}
    \label{tab:env_overview}
\end{table}

\vspace{-.25cm}

\paragraph{Evaluation Teammates}

For each environment, JaxAHT provides a standardized evaluation teammate suite consisting of hand-designed heuristic teammates, algorithmic teammates trained with IPPO, BRDiv, LBRDiv, and CoMeDi, and teammates derived from human gameplay data.
Each teammate is accompanied by an empirical BR reference, used to normalize returns as described below. \looseness=-1 %

The heuristic teammates are hand-designed policies that induce distinct coordination demands across environments. In LBF, they target foods according to fixed-ordering, greedy, or wait-for-partner rules; in Overcooked-v1, they implement role-based strategies; and in Hanabi, they consist of the Walton-Rivers heuristic suite~\citep{waltonrivers2017} and SmartBot~\citep{smartbot2019}. Detailed descriptions of the heuristic teammate suite are provided in App.~\ref{app:teammate_descriptions}.

On Overcooked, the human-data-derived policies are trained via behavior cloning~\citep{bainFrameworkBehaviouralCloning1999} using the human gameplay data released by \citet{carroll_utility_2019}. 
For LBF, we collect a human gameplay dataset that we open-source with this work (collection procedure described in App.~\ref{app:human_data_lbf}), and use human-regularized RL~\citep{jacob2022modelingstrong,cornelisseHumancompatibleDrivingPartners2024} to derive a performant, human-like policy.
For Hanabi, we refer users to the gold-standard, API-gated human proxies of AH2AC2~\citep{dizdarevic2025ah2ac2}, as described in Sec.~\ref{sec:related_work}.

\paragraph{Evaluation Metrics}
\label{sec:eval_metrics}
Raw returns are not comparable across teammates of differing competence, as a low team return could reflect either a poor AHT agent or simply a weak teammate.
To address this, \citet{wang2024zsceval} proposed the \emph{BR-normalized return} metric, which replaces $J(\pi_e,\boldsymbol{\pi}_{-e})$ in Eq.~(\ref{eq:aht_objective}) with $J(\pi_e,\boldsymbol{\pi}_{-e})/J(\mathrm{BR}(\boldsymbol{\pi}_{-e}),\boldsymbol{\pi}_{-e})$.
JaxAHT adopts this best-practice recommendation.
As exact BRs are intractable, the denominator is estimated using the empirical BR reference provided for each evaluation teammate, which serves as a per-teammate upper bound (see Sec.~\ref{sec:exp:challenges_br} for a discussion of the limitations of this estimate).
Normalized returns are aggregated over teammates and reported with 95\% bootstrapped confidence intervals, stratified across teammates and computed using \texttt{rliable}~\citep{agarwal_deep_2021}.

\vspace{-.2cm}
\section{Experimental Studies}
\label{sec:experiments}
\vspace{-.2cm}
\begin{wraptable}[10]{r}{0.5\textwidth} %
    \vspace{-0.5cm}
    \centering
    \small
    \begin{tabular}{lccc}
        \hline
        Algorithm & Seeds & PyTorch & JAX \\
        \hline
        LIAM  & 1   & 40m 27s & \textbf{08m 22s} \\
        BRDiv & 1   & 44m 18s & \textbf{01m 46s}  \\
        \hline
        LIAM  & 3   & 121m 21s*  & 09m 10s \\
        LIAM  & 10  & 404m 30s* & 12m 17s \\
        LIAM  & 300 & 202h 15m* & 126m \\
        \hline
    \end{tabular}
    \caption{\textbf{Wall-clock time comparison on LBF 7x7.} JaxAHT achieves ${\sim}5$x (LIAM) and ${\sim}25$x (BRDiv) single-seed speedups over PyTorch, and 95x for LIAM across 300 seeds. Asterisks indicate runtimes for sequentially running PyTorch implementations.}
    \label{tab:wallclock_pytorch_jax}
\end{wraptable}
This section addresses three questions using JaxAHT: (1) What is the speedup of JaxAHT over conventional PyTorch frameworks? (2) What is the quality and diversity of the JaxAHT evaluation teammates? (3) How well do agents trained by the library's algorithms coordinate with the evaluation teammates? 
It closes by discussing the finding that BR returns are often underestimated by training PPO against a single teammate, and the implications for AHT algorithms that rely on BR estimation (Sec.~\ref{sec:exp:challenges_br}).
\vspace{-.2cm}
\subsection{Wall Clock Time Comparison}
\vspace{-.2cm}
We compare the wall-clock time of representative teammate generation and ego agent algorithms in JaxAHT against their original PyTorch implementations (Table~\ref{tab:wallclock_pytorch_jax}).
All runs use a single Nvidia A6000 GPU (48 GB VRAM), 8 environment runners, and 3 million timesteps, and all reported figures include JIT compilation time.
Unlike similar JAX-based libraries (e.g., JaxMARL~\citep{rutherford2024jaxmarl}), which report speedups from benchmarking environment step speed alone with several thousand parallel environments, we benchmark full algorithms with fewer environments and many seeds.
This design reflects a more conventional usage scenario, as previous AHT algorithms were not designed with large-scale, JAX-based data collection capabilities in mind.
Note that as GPU memory is the ultimate constraint, parallelizing over environments trades off against parallelizing over seeds (or over populations, as in teammate generation). \looseness=-1 %

Table \ref{tab:wallclock_pytorch_jax} demonstrates that in the single-seed scenario, JAX yields a $5\text{x}$ runtime improvement over PyTorch for LIAM and a $25\text{x}$ improvement for BRDiv.
Training 300 seeds of LIAM on a single GPU yields an estimated $95\text{x}$ speedup over the corresponding PyTorch implementation.
The speedup occurs because a standard PyTorch implementation's ability to parallelize computation is bounded by the number of CPU threads, while our JAX-based implementations can leverage GPU parallelization. \looseness=-1 %

\vspace{-.35cm}
\subsection{Evaluation Teammate Quality and Diversity}
\label{sec:exp:teammates}
\vspace{-.2cm}

The JaxAHT teammates are generated by applying RL, manually programming heuristic strategies, and by deriving policies from human gameplay data. The quality of the JaxAHT teammate suite is assessed by reporting the estimated BR return achievable against each teammate, while diversity is assessed via the cross-play (XP) matrix of returns between teammate-BR pairs and by visualizing latent embeddings of behavioral trajectories.

\begin{figure}[!t]
    \centering
    \includegraphics[width=\linewidth]{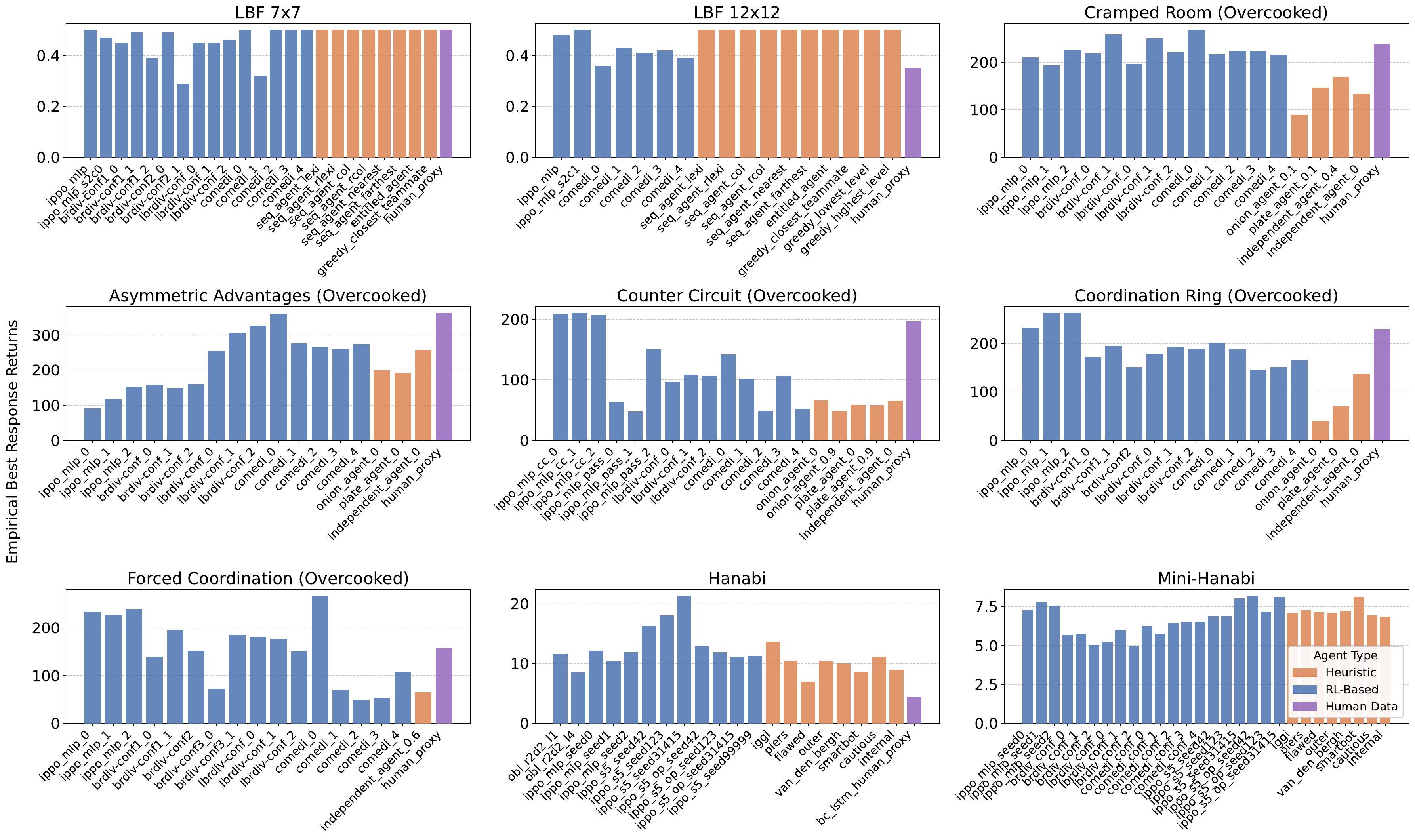}
    \caption{\small \textbf{BR return estimates for JaxAHT evaluation teammates.} Each bar is the estimated BR return for an evaluation teammate, colored by teammate type. Estimates are the maximum return obtained against that teammate by any trained BR policy in the cross-play matrix, or the known optimal return for certain LBF heuristics. Note that the maximum return on LBF is 0.5, on Hanabi is 25, and on Mini Hanabi is 8. For Overcooked-v1 tasks, agents are awarded 20 points per delivered dish, with no analytic maximum.\looseness=-1}
    \label{fig:br_returns_from_config}
\end{figure}

\vspace{-.2cm}
\paragraph{Teammate Quality} 
For most teammates, BR returns are estimated by training a PPO agent against each teammate, evaluating all teammates against all BRs to form an XP matrix, and taking the maximum return observed against each teammate.
For certain LBF heuristic teammates, the BR return is known exactly, since an optimal partner can be programmed.
The full procedure is reported in App.~\ref{app:br_upper_bounds}. \looseness=-1 %
As shown in Fig.~\ref{fig:br_returns_from_config}, JaxAHT evaluation teammates achieve a wide range of returns, reflecting a range of coordination difficulty provided by each teammate, with  some teammates achieving near-optimal returns. 

\vspace{-.2cm}
\paragraph{Return-Based Diversity}
Following prior work~\citep{agapiou2022meltingpot, rahman2023brdiv,nekoei2023fewshotcoordination}, we use the XP matrix to assess return-based diversity of teammates.
If teammates are high-quality and behaviorally diverse, each should require a distinct response strategy, yielding an XP matrix with high diagonal and lower off-diagonal entries.
App. Figs.~\ref{fig:lbf-xp-return}--\ref{fig:overcooked-xp} display the XP matrices for LBF, Hanabi, and Overcooked domains.
Diagonal entries are generally stronger, indicating that high returns across the suite require an AHT agent to model distinct response strategies, although some off-diagonal blocks retain high returns, suggesting partial behavioral overlap or transferability across response strategies, especially within agent types (e.g., PPO).

\begin{wrapfigure}[15]{r}{0.5\textwidth}
    \vspace{-0.7cm}
    \centering
    \includegraphics[width=\linewidth]{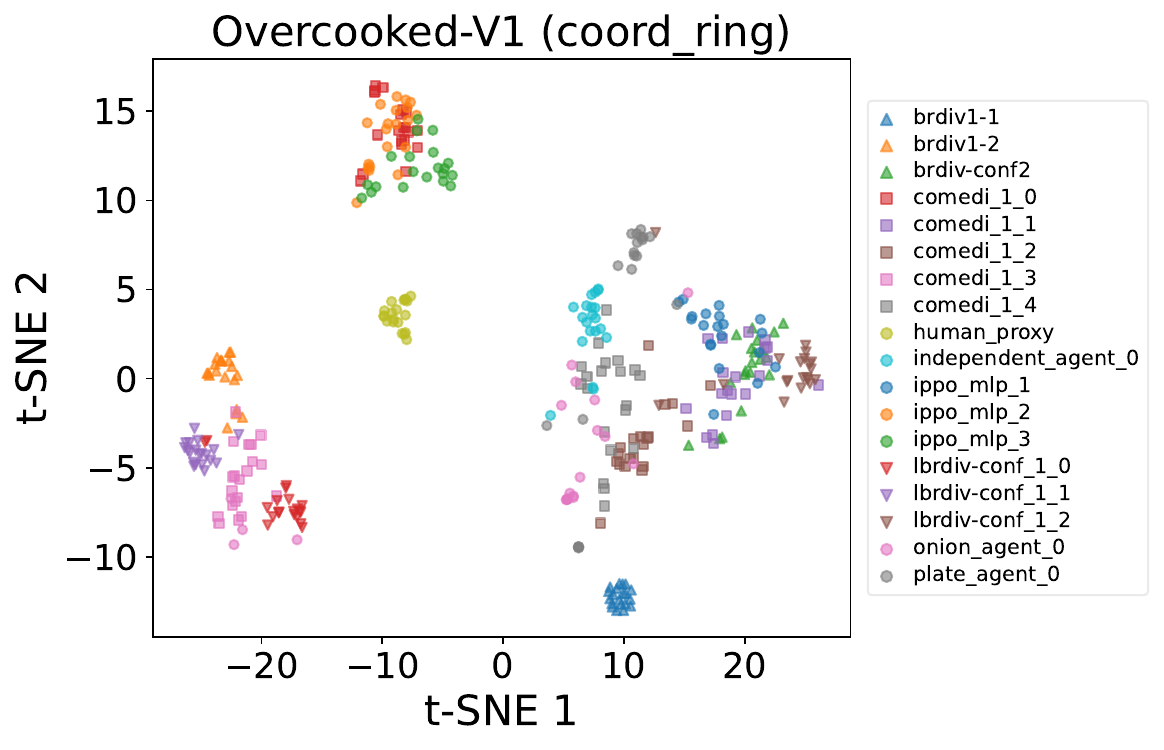}
    \caption{\small\textbf{Representations of Behavior Trajectories for Evaluation Teammates.} Each point reflects a single embedded  behavior trajectory of a teammate-BR pair. Teammate-BR pairs are meaningfully distinct.}
    \label{fig:overcooked-tsne-visualization}
\end{wrapfigure}

\vspace{-.2cm}
\paragraph{Behavioral Diversity}
The previous return-based analysis determines whether teammates require distinct BRs, but not whether their behavioral trajectories differ. Two teammates with separate BRs may produce similar trajectories, suggesting that the BRs differ only because they are brittle to small perturbations in teammate behavior, rather than because the teammates exhibit distinct conventions.
To assess behavioral diversity directly, we train an LSTM classifier on the joint observation--action sequences of teammate--BR pairs to predict pair identity, and embed its latent representations with t-SNE~\citep{vandermaaten2008tsne}.
Fig.~\ref{fig:overcooked-tsne-visualization} shows the embeddings for behavior trajectories in Overcooked Coordination Ring, where each point reflects the trajectory of an evaluation teammate paired with its own BR.
Teammate-BR pairs occupy largely non-overlapping regions, confirming that the evaluation set achieves behavior-level diversity.
See App.~\ref{app:tsne_details} for the full procedure and LBF results. \looseness=-1 %

\subsection{Benchmarking AHT Algorithm Performance}
\label{sec:exp:benchmark}

To illustrate the studies JaxAHT enables, we benchmark all AHT algorithms in JaxAHT across the suite. A systematic, controlled comparison across algorithms and environments at this scale has not previously been feasible, and both the results and the methodology are intended to motivate further investigation.
The benchmark uses a representative task subset: LBF 7x7 and 12x12, Overcooked Cramped Room (CR) and Coordination Ring (CoR), and Mini Hanabi.\footnote{Counter Circuit is excluded due to similarity to CoR, while Asymmetric Advantages and Forced Coordination are excluded because \citet{wang2024zsceval} found they do not substantially differentiate AHT algorithm performance.
Full Hanabi is excluded as its computational cost is prohibitive for the hyperparameter sweep procedure (IPPO self-play typically requires $\sim\!10^9$ training steps).}
We outline the procedure here, then present the hyperparameter sweep and benchmark results.
See App.~\ref{app:hp_sweep} and \ref{app:benchmark_procedure} for full details.

\vspace{-.2cm}
\paragraph{Procedure} 
For each algorithm--task pair, 140 hyperparameter configurations are randomly sampled (or all configurations searched, if fewer configurations exist).
The best configuration is selected by normalized mean return on a validation set of teammates that are disjoint from both the training and evaluation teammates, so that no configuration is benchmarked on teammates it was tuned against.
Training timesteps are strictly controlled for both the sweep and the benchmark.
The benchmark budget is 195M timesteps for teammate generation and unified methods and 30M for ego agent methods on LBF 7x7 and 12x12, Mini Hanabi, and CR. The budget is increased to 390M and 60M respectively on CoR, where methods converge more slowly.
Budgets were selected so that even the population-based methods receive sufficient data for optimization.
The benchmark study compares methods by mean normalized returns and 95\% stratified bootstrapped confidence intervals, as discussed in Sec.~\ref{sec:jaxaht}. \looseness=-1 %

\begin{figure}[!t]
    \centering
    \begin{subfigure}[b]{\linewidth}
        \centering
        \includegraphics[width=0.9\linewidth, trim=0 0 0 30pt, clip]{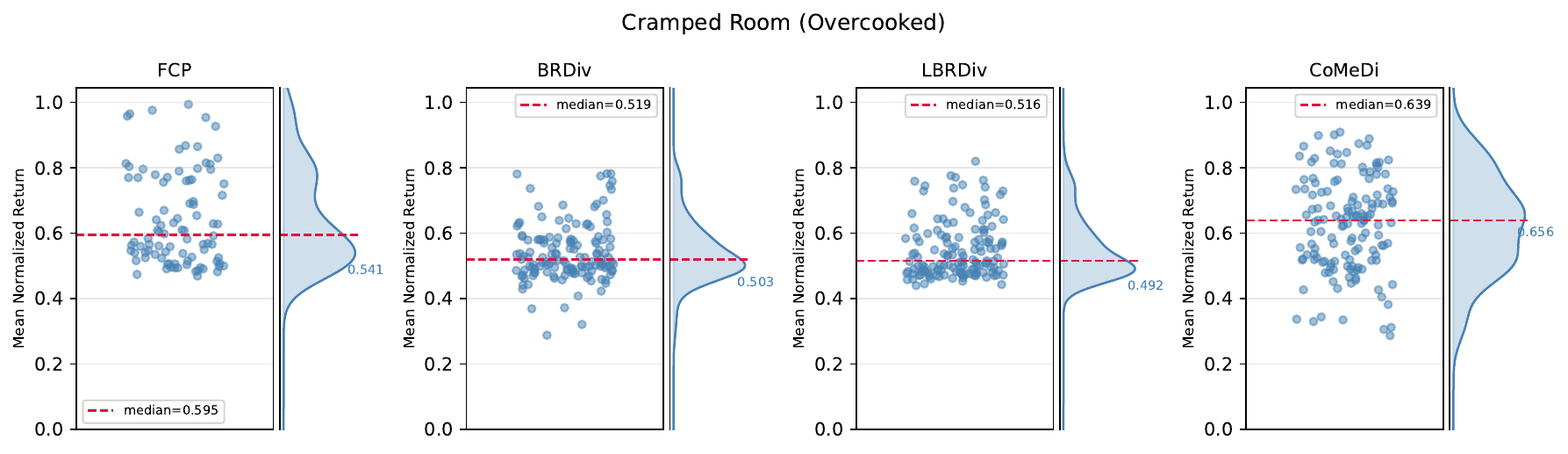}
        \caption{Teammate generation methods.}
    \end{subfigure}
    \\[0.5em]
    \begin{subfigure}[b]{\linewidth}
        \centering
        \includegraphics[width=0.7\linewidth, trim=0 0 0 30pt, clip]{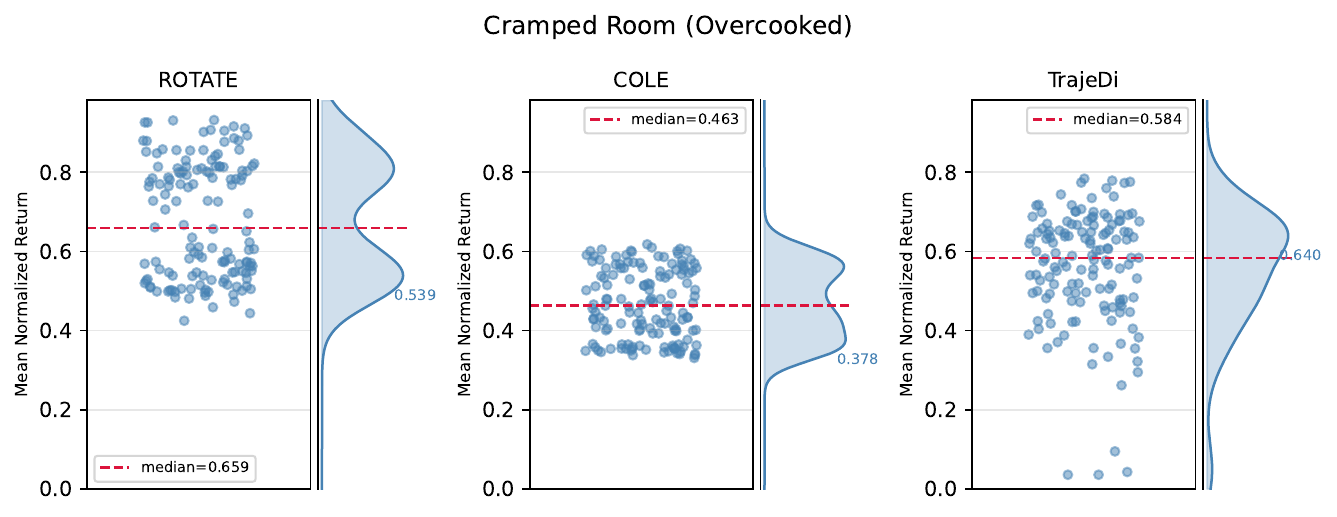}
        \caption{Unified methods.}
    \end{subfigure}
    \caption{\textbf{Hyperparameter sweep return distributions on Cramped Room.} Each point represents the performance of a distinct hyperparameter configuration on a validation set of teammates. Best performing configurations are selected for the benchmark study.}    
    \label{fig:sweep_dist_unified_cramped_room}
\end{figure}

\vspace{-.2cm}
\paragraph{Hyperparameter Sweep Results}
Fig.~\ref{fig:sweep_dist_unified_cramped_room} displays the normalized return distributions over swept hyperparameters for all teammate generation and unified AHT methods on CR.  
Those for the ego agent methods and for LBF 12x12 are displayed in App. Figs.~\ref{fig:sweep_dist_ego_cramped_room} and \ref{fig:sweep_dist_lbf}. 
We swept each algorithm's unique hyperparameters, plus the learning rate, entropy coefficient, and clip parameter of the base RL algorithm (PPO). 
The remaining base hyperparameters were fixed to the same values for all methods. Configurations were sampled by random search, except for FCP (exhaustive grid search over its 90 configurations) and CoR (Bayesian search); see App.~\ref{app:hp_sweep}.

\begin{figure}[t]
    \centering
    \begin{subfigure}{0.57\linewidth}
        \includegraphics[width=\linewidth]{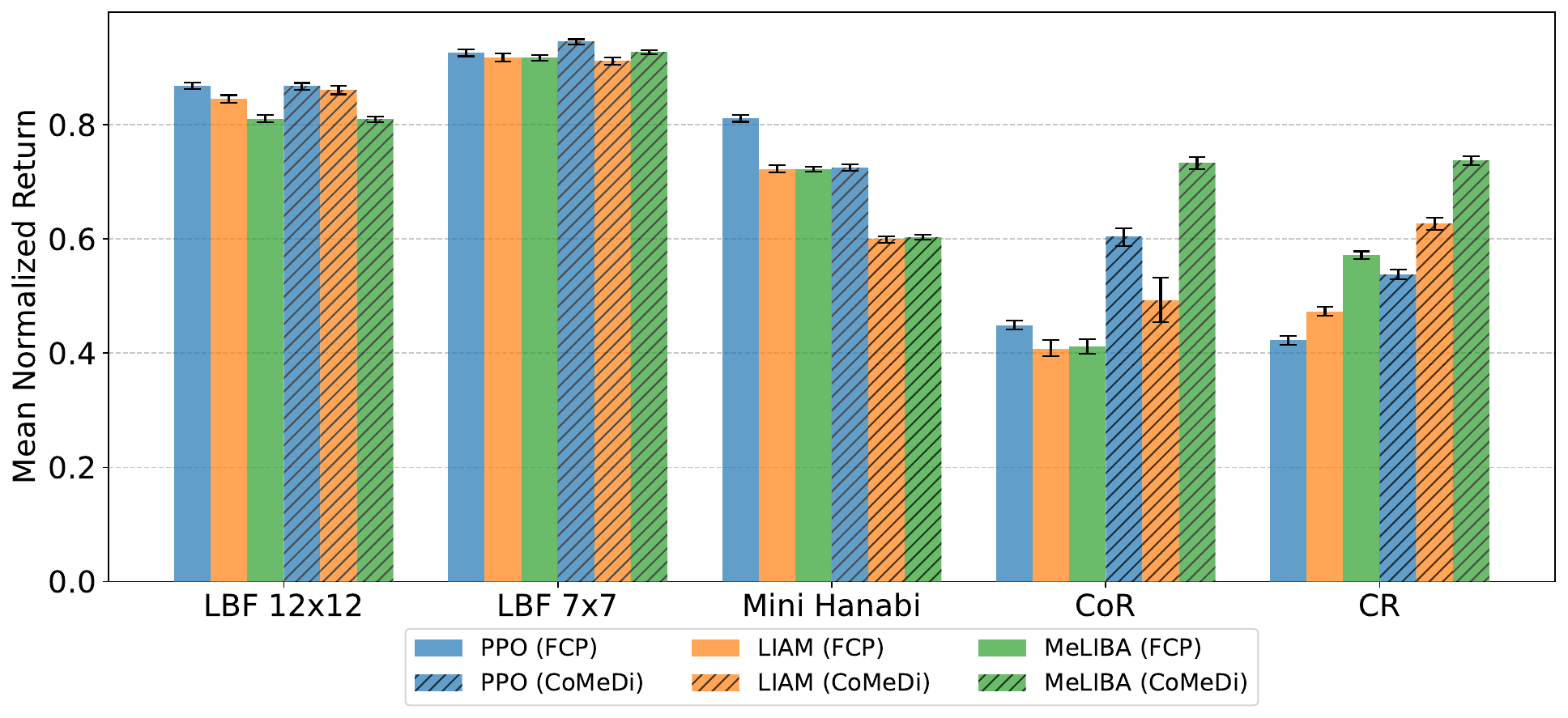}
        \caption{\small Ego agent methods.}
        \label{fig:ego_benchmark}
    \end{subfigure}
    \hfill
    \begin{subfigure}{0.42\linewidth}
    \includegraphics[width=0.97\linewidth]{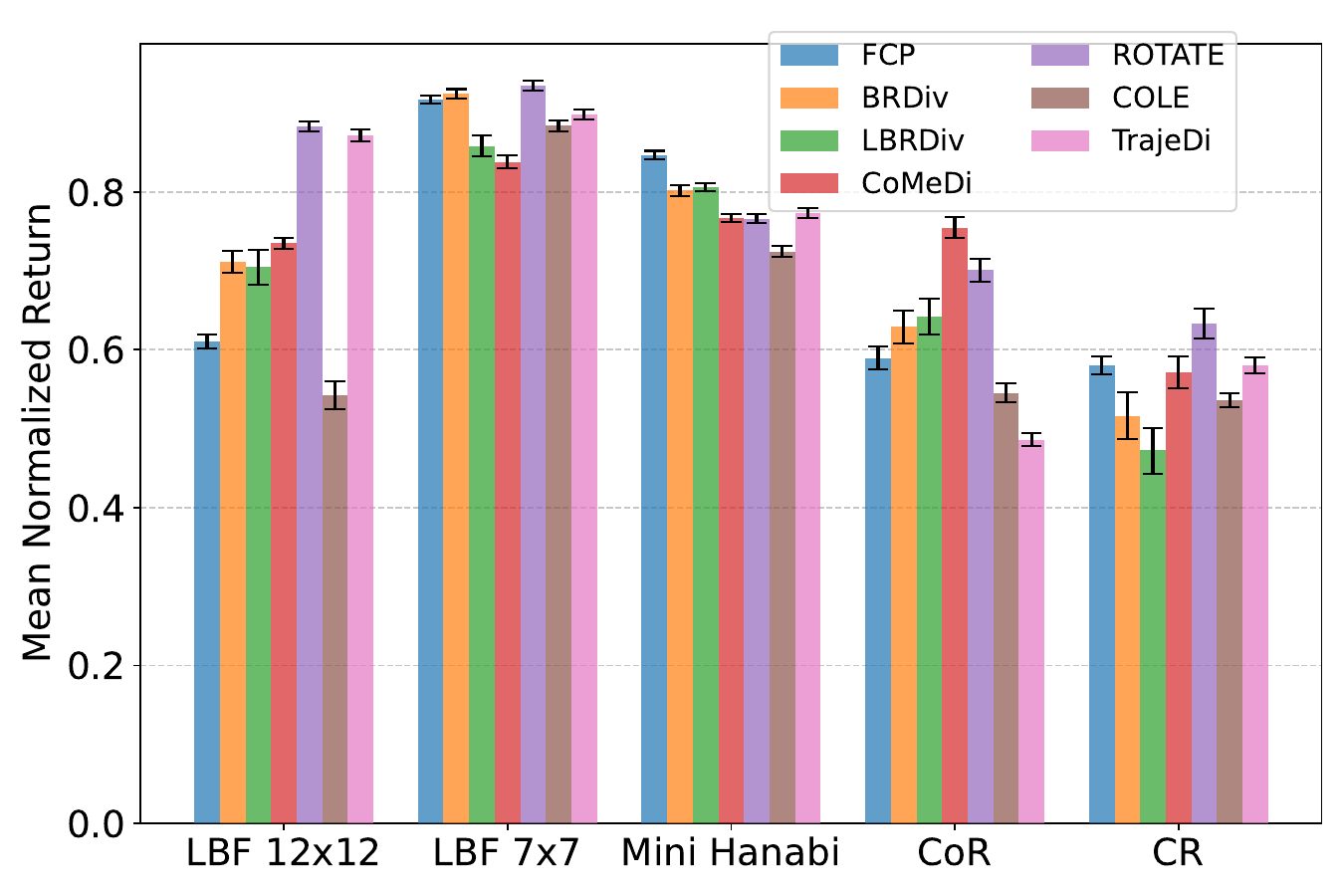}
        \caption{\small Teammate generation and unified methods.}
        \label{fig:unified_benchmark}
    \end{subfigure}
    \caption{\textbf{AHT Algorithm Benchmark Results.} (Left) Ego training methods with agent modeling (LIAM, MeLIBA) outperform PPO only in Overcooked environments, which have role-based conventions, and in the presence of diverse teammate populations (CoMeDi). (Right) No single AHT method dominates on all tasks; the best overall are ROTATE, FCP, and CoMeDi.}
    \label{fig:benchmark_results}
\end{figure}

Hyperparameters induce a wide range of performance, underscoring the need for compute-controlled tuning when comparing algorithms.
Such tuning is infeasible at academic-scale compute under existing PyTorch implementations, but is now possible with JaxAHT.
The optimized hyperparameters are released with the library as a starting point for researchers and to enable fairer future comparisons. 
\vspace{-.3cm}
\paragraph{Benchmark Study Results}
AHT algorithm performance on the JaxAHT evaluation teammate suite is compared using the best discovered hyperparameters in Fig.~\ref{fig:benchmark_results}, measured over 5 seeds.
Our analysis first addresses ego agent methods, then teammate generation and unified methods. Learning curves, losses, and key statistics for all algorithms are in App. Fig.~\ref{fig:train_loss_curves}.

Fig.~\ref{fig:ego_benchmark} compares PPO, LIAM, and MeLIBA across training populations generated by FCP and CoMeDi. 
The FCP population consists of seeds and intermediate checkpoints of IPPO resulting in  a range of skill levels, while CoMeDi agents maximize an adversarial diversity objective designed to combat self-sabotage~\citep{strouse2021fcp, sarkar2023comedi}, yielding more varied cooperative conventions.
JaxAHT's unified support for both stages makes it possible to study how ego agent performance depends on training population diversity, a line of inquiry previously hindered by incompatible implementations.

LIAM is expected to outperform PPO via explicit autoencoder-based agent modeling, and MeLIBA to improve further via meta-learning. However, on both LBF tasks, all three perform similarly, indicating little-to-no benefit from agent modeling, and on Mini Hanabi, LIAM and MeLIBA achieve markedly lower returns than PPO.
The opposite trend emerges on Overcooked, where MeLIBA substantially outperforms PPO and LIAM for both training populations, with LIAM outperforming PPO on CR.
We attribute the split to the role structure of the tasks: unlike LBF and Mini Hanabi, Overcooked layouts require partners to occupy complementary roles, so the ego agent's BR strategy varies sharply with teammate identity.
We hypothesize that representing BRs to such role-based behaviors (e.g., transporting onions vs. delivering dishes) is difficult without latent representations of teammate behavior~\citep{mieczkowskiPredictingMultiAgentSpecialization2025}, allowing explicit agent modeling to outperform PPO.
The absence of a benefit on LBF and Mini Hanabi is consistent with the finding that recurrent PPO agents implicitly model teammates when the task demands it \citep{monwilliamsPartnerModellingEmerges2025}.
If implicit modeling suffices, an explicit auxiliary objective adds no information and may instead \textit{conflict} with return maximization~\citep{liuConflictaverseGradientDescent2021}, explaining the slightly lower returns on LBF and markedly lower returns on Mini Hanabi.
Designing agent modeling objectives that avoid such interference while preserving benefits in role-based tasks remains an open direction. 

Fig.~\ref{fig:unified_benchmark} displays the performance of the teammate generation and unified methods. While no single method performs best across all scenarios, ROTATE, FCP, and CoMeDi are the strongest overall, each performing best on at least one task. 
COLE is frequently the weakest, consistent with \citet{wang2024zsceval}. 
Per-teammate-type analysis (App. Fig.~\ref{fig:by_agent_type_teamgen_unified}) shows COLE performs best against IPPO teammates, suggesting its population closely resembles IPPO. 
Although COLE upweights the most cooperatively incompatible teammates during training, its diversity remains limited to conventions discoverable through return maximization, likely explaining why it underperforms FCP despite a similar population structure.
FCP and TrajeDi are surprisingly competitive given that their limitations motivated the adversarial diversity methods that followed. 
Prior comparisons may have been less compute-controlled than ours.

Considering the average return over algorithms, the Overcooked variants and Mini Hanabi pose greater learning challenges than the LBF variants. Additionally, teammate generation methods perform notably worse on LBF 12x12 than on LBF 7x7, whereas unified and ego agent methods perform comparably on both, suggesting that the former, population-based methods scale less well to larger state spaces despite substantially larger training budgets. 

\vspace{-.2cm}
\subsection{Challenges of Best Response Return Estimation}
\label{sec:exp:challenges_br}
\vspace{-.2cm}

Despite the two-stage BR estimation procedure of Sec.~\ref{sec:exp:teammates}, AHT agents from the benchmark study sometimes outperformed the PPO-derived BR return estimates (App. Fig. \ref{fig:br_return_delta})\footnote{For the benchmark study, the BR return estimates used to normalize returns were revised upwards with the best seen returns per-teammate, to minimize bias arising from BR return underestimation.}. 
We attribute this to benefits of transfer and generalization: AHT agents are trained against diverse populations, whereas the PPO BR is trained against a single teammate that may be hard to learn against.

There are a variety of implications for AHT research, as some AHT algorithms are developed on the premise of being able to use RL algorithms to estimate the BR policy to an AHT agent~\citep{rahman2023brdiv,rahman2024lbrdiv,li2023cole}, or leverage BR policies as part of the procedure~\citep{barrett2017makingfriends}.
First, evaluations of such algorithms should consider, theoretically or empirically, their sensitivity to BR estimation error.
Second, BR estimation may be improved by leveraging pretrained, generally capable agents, an avenue that to our knowledge has not previously been considered for AHT.

\vspace{-.2cm}
\section{Discussion and Conclusion}
\label{sec:conclusion}
\vspace{-.2cm}
This paper introduces JaxAHT, the first open-source, JAX-based library for the full AHT research lifecycle, 
achieving approximately $95\times$ wall-clock speedup over PyTorch counterparts and making compute-controlled, large-scale AHT studies tractable at academic scale.
Alongside the library, we contribute a diverse suite of validated evaluation teammates across LBF, Overcooked, and Hanabi. 
We also conduct the first benchmark study comparing various teammate generation and ego agent methods under controlled compute budgets. 
Our results show that no single algorithm performs best, and that agent-modeling-based AHT algorithms  help primarily in role-based tasks, and sometimes harm learning compared to a non-agent-modeling baseline.
We also find that BR returns may be systematically underestimated by standard single-agent PPO training, with direct implications for algorithms relying on BR estimation.

JaxAHT opens several directions supporting future work. 
The unified framework enables joint study of teammate generation and ego agent learning, an interaction that has been historically difficult to investigate due to incompatible implementations.
The released hyperparameters and evaluation teammates provide a foundation for fairer comparisons.
Finally, more capable BR estimation based on strong cooperative priors (such as provided by AHT agents) is a promising and underexplored direction. \looseness=-1 %

We identify two limitations.
First, we focus on existing AHT benchmark tasks rather than designing new ones. 
Environments with richer state spaces, continuous control, and more complex role structure are equally important for AHT progress, and represent a complementary direction.
Second, the benchmark study was conducted on a subset of tasks due to compute limits. 
Thus, extending the study to cover all tasks is a direction of future work. 
\looseness=-1 %

\newpage
\section*{AI Use Statement}
This work used generative AI tools to assist with implementing experimental code, creating figures, and polishing the writing of this manuscript.
We have not used generative AI tools for synthetic data generation, mathematical formulations or proofs, research methodology design, or results analysis. 
All ideas, arguments, and original text were produced by the authors.
We have reviewed all AI-assisted work and take full responsibility for the final content of the paper. 

\section*{Ethics Statement}
The LBF human gameplay dataset was collected with IRB approval. As detailed in App.  App.~\ref{app:human_data_lbf}, participation was voluntary, and participants were compensated through Prolific at 8--9 USD per hour.
All third-party code, data, and model weights are used in accordance with their licenses (App.~\ref{app:attributions}).

\section*{Reproducibility Statement}
The JaxAHT library, evaluation teammate and BR policies, and LBF human gameplay data have been open-sourced at the links in App.~\ref{app:code_data}. Code to reproduce the experiments and the tuned hyperparameters are released with the library.
The hyperparameter sweep and benchmark procedures, including timestep budgets, are described in Sec.~\ref{sec:exp:benchmark} and App.~\ref{app:benchmark_details}, BR estimation in App.~\ref{app:br_upper_bounds}, and compute infrastructure in App.~\ref{app:compute}.

\bibliography{references}
\bibliographystyle{iclr2027_conference}

\newpage
\raggedbottom
\appendix

\section{Code and Data}
\label{app:code_data}

\ifpreprint
The library is open-sourced on GitHub, while the datasets of evaluation teammates, best responses, and LBF human gameplay data are released on Hugging Face.

\begin{itemize}[noitemsep,leftmargin=*]
    \item \textbf{Code:} \url{https://github.com/LARG/jax-aht}
    \item \textbf{Evaluation Teammate Policies:} \newline\url{https://huggingface.co/datasets/jaxaht/eval-teammates}
    \item \textbf{Best Response Policies:} \newline\url{https://huggingface.co/datasets/jaxaht/eval-teammates-br}
    \item \textbf{LBF Human Gameplay Data:} \newline\url{https://huggingface.co/datasets/jaxaht/lbf-human-data}
\end{itemize}
\else
Anonymized links are provided below.

\begin{itemize}[noitemsep,leftmargin=*]
    \item \textbf{Code:} \url{https://anonymous.4open.science/r/jax-aht-F125}
    \item \textbf{Evaluation Teammate Policies:} \newline\url{https://anonymous-hf.com/a/ai5usqubdcox/}
    \item \textbf{Best Response Policies: } \newline\url{https://anonymous-hf.com/a/k6mig3aw8gnf/}
    \item \textbf{LBF Human Gameplay Data:} \newline\url{https://anonymous-hf.com/a/arv719qpjjmf/}
\end{itemize}
\fi

\section{Best-Response Policies and Upper Bound Estimation}
\label{app:br_upper_bounds}
To assess behavioral diversity among the evaluation teammates, we construct a cross-play (XP) matrix between each teammate and a corresponding best-response (BR) policy.

\paragraph{Training protocol.}
BR policies are trained using the \texttt{ppo\_br} implementation, fixing one teammate policy per run and optimizing an ego policy against that teammate alone. Each BR run uses a budget of 10M timesteps.

The PPO-BR ego policies use a recurrent S5 actor--critic architecture~\citep{lu_s5_2023}. Observations are encoded through dense layers and passed to an S5 sequence model with model dimension 128 and SSM state size 128. Separate MLP heads (three hidden layers of 1024 units each) produce a masked categorical action distribution and a scalar value estimate, respectively.

\begin{figure}[ht]
    \centering
    \begin{subfigure}[t]{0.48\linewidth}
        \centering
        \includegraphics[width=\linewidth]{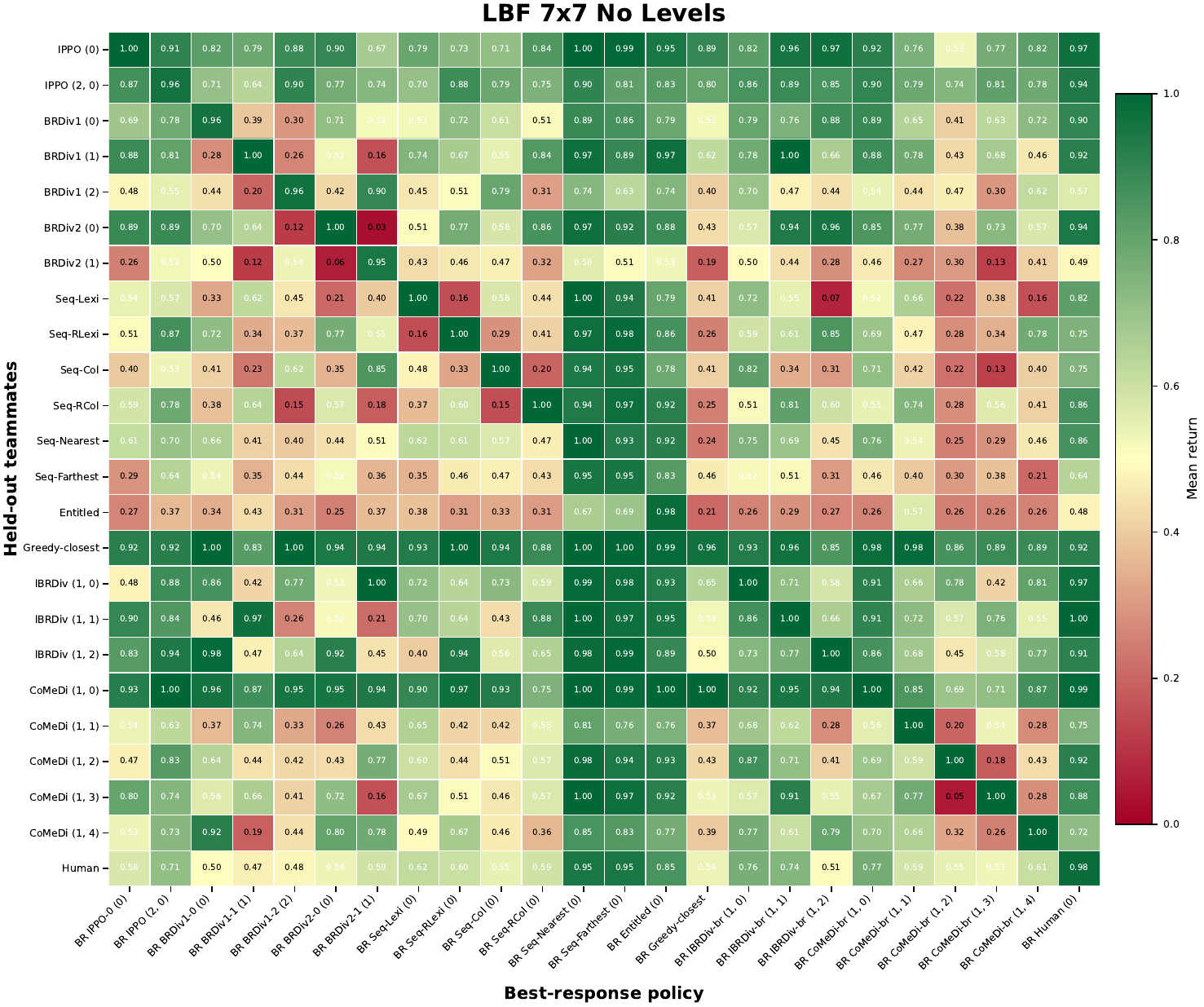}
        \caption{LBF 7$\times$7.}
        \label{fig:lbf-7x7-xp-return}
    \end{subfigure}
    \hfill
    \begin{subfigure}[t]{0.48\linewidth}
        \centering
        \includegraphics[width=\linewidth]{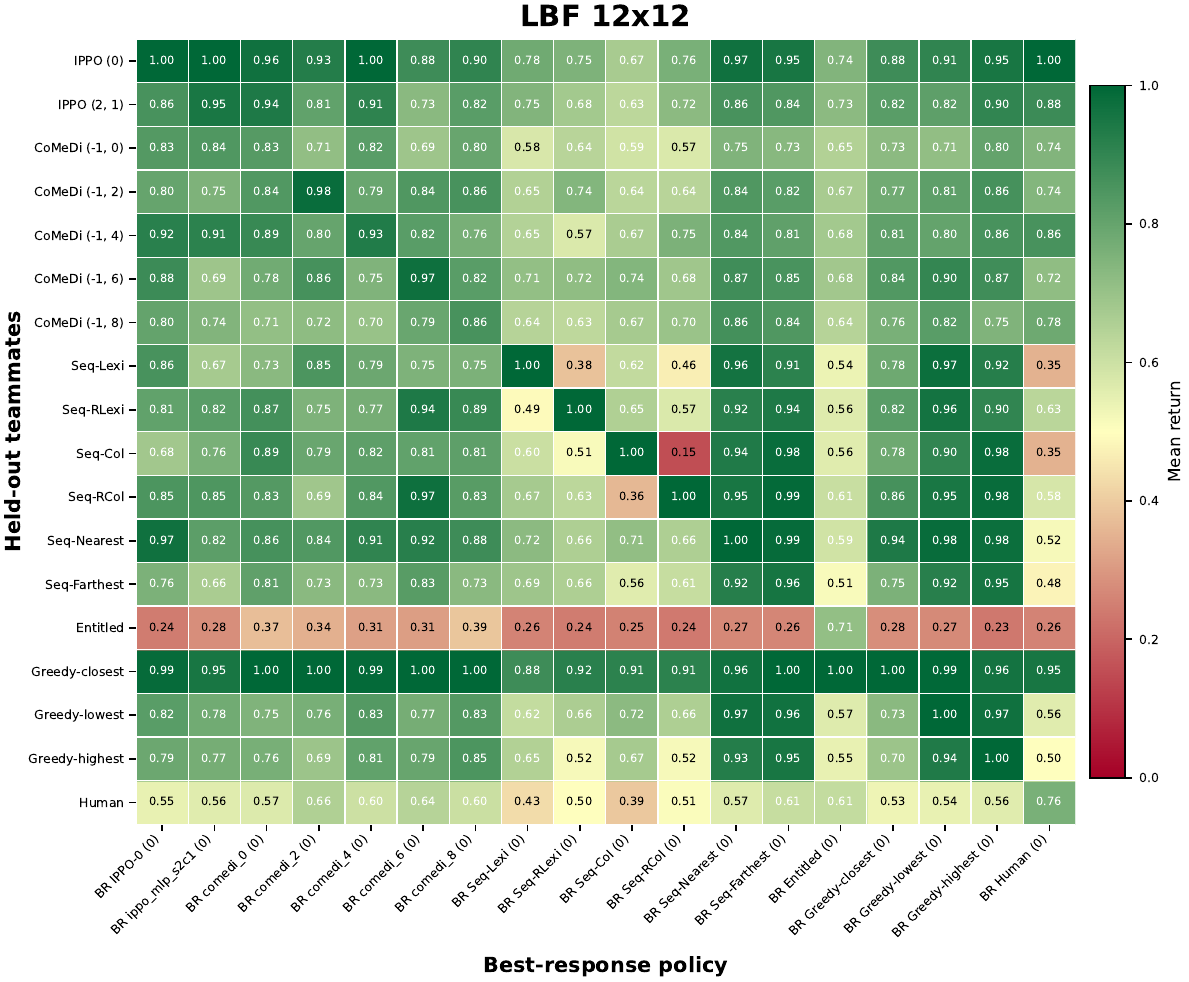}
        \caption{LBF 12$\times$12.}
        \label{fig:lbf-12x12-xp-return}
    \end{subfigure}
    \caption{\textbf{LBF cross-play matrices using normalized return.} Rows are evaluation teammates and columns are best-response policies.}
    \label{fig:lbf-xp-return}
\end{figure}

\begin{figure}[ht]
    \centering
    \begin{subfigure}[t]{0.48\textwidth}
    \includegraphics[width=\textwidth]{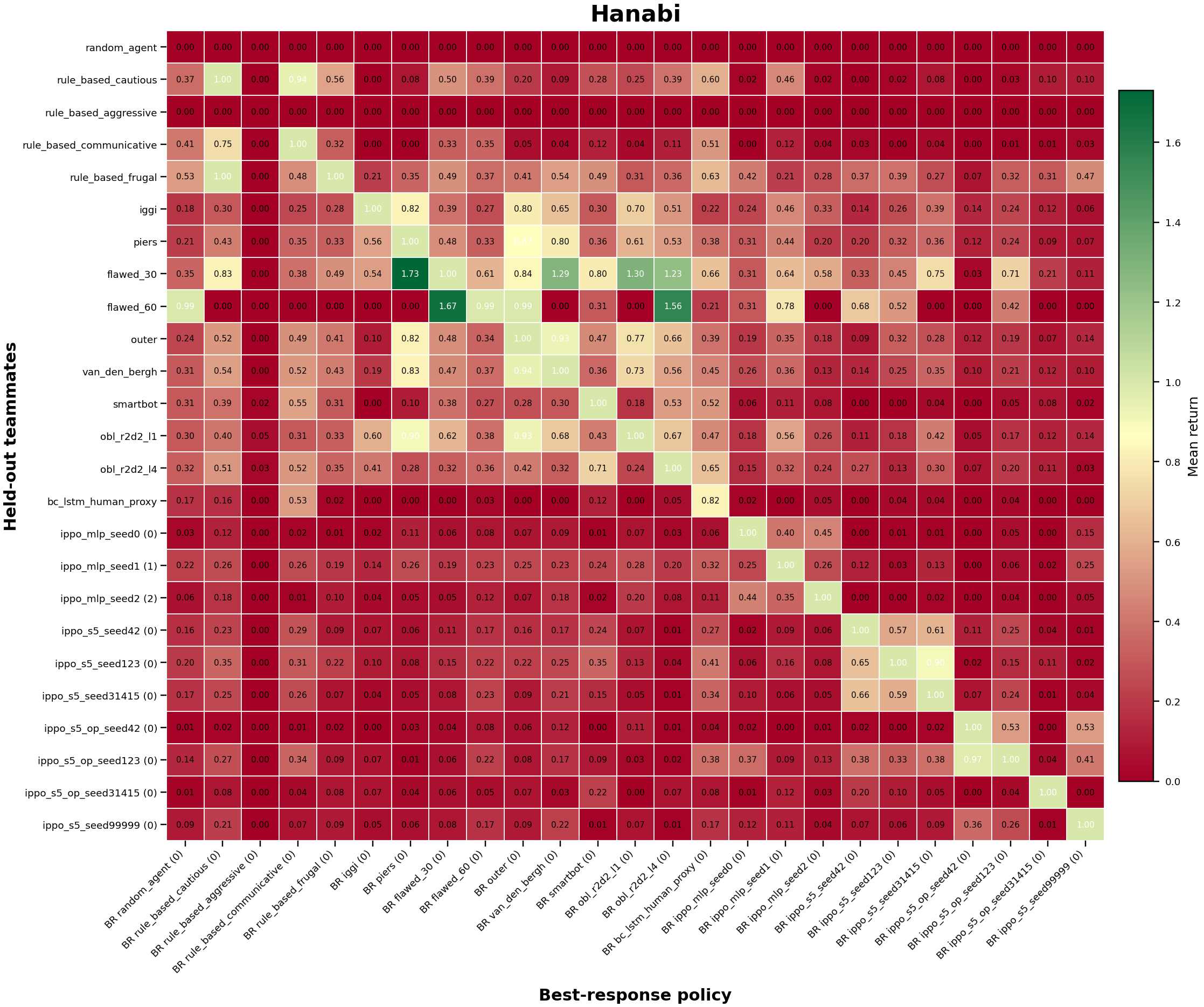}
    \caption{Hanabi.}
    \label{fig:full-hanabi-xp} 
    \end{subfigure} 
    \begin{subfigure}[t]{0.48\textwidth} 
    \centering
    \includegraphics[width=\textwidth]{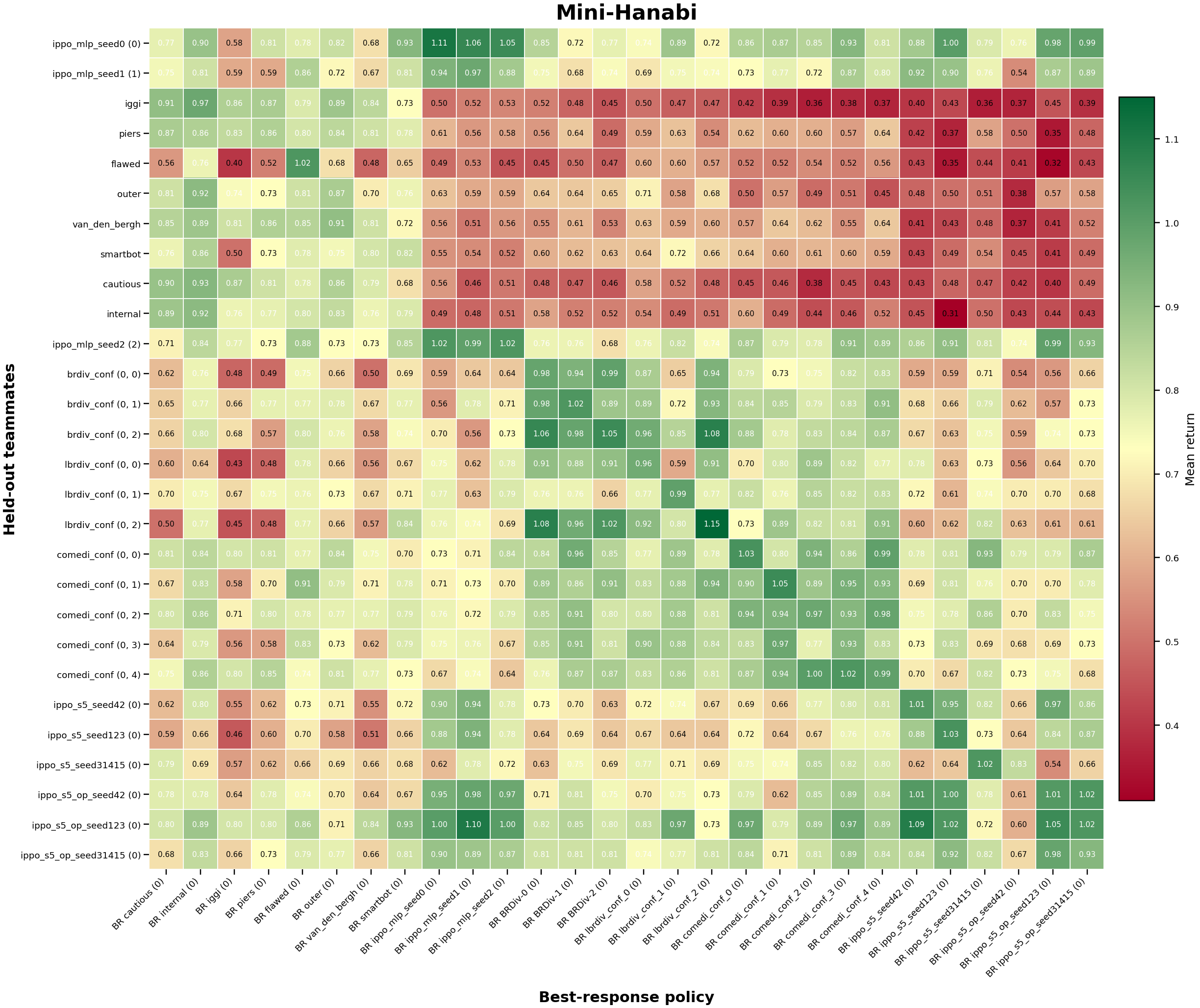}
    \caption{Mini Hanabi.}
    \label{fig:mini_hanabi_xp}
    \end{subfigure} 
    \caption{\textbf{Hanabi cross-play matrices using normalized return.} Rows are evaluation teammates and columns are best-response policies.}
    \label{fig:hanabi_xp}
\end{figure}

\begin{figure}[p]
    \centering
    \begin{subfigure}[t]{0.48\linewidth}
        \centering
        \includegraphics[width=\linewidth]{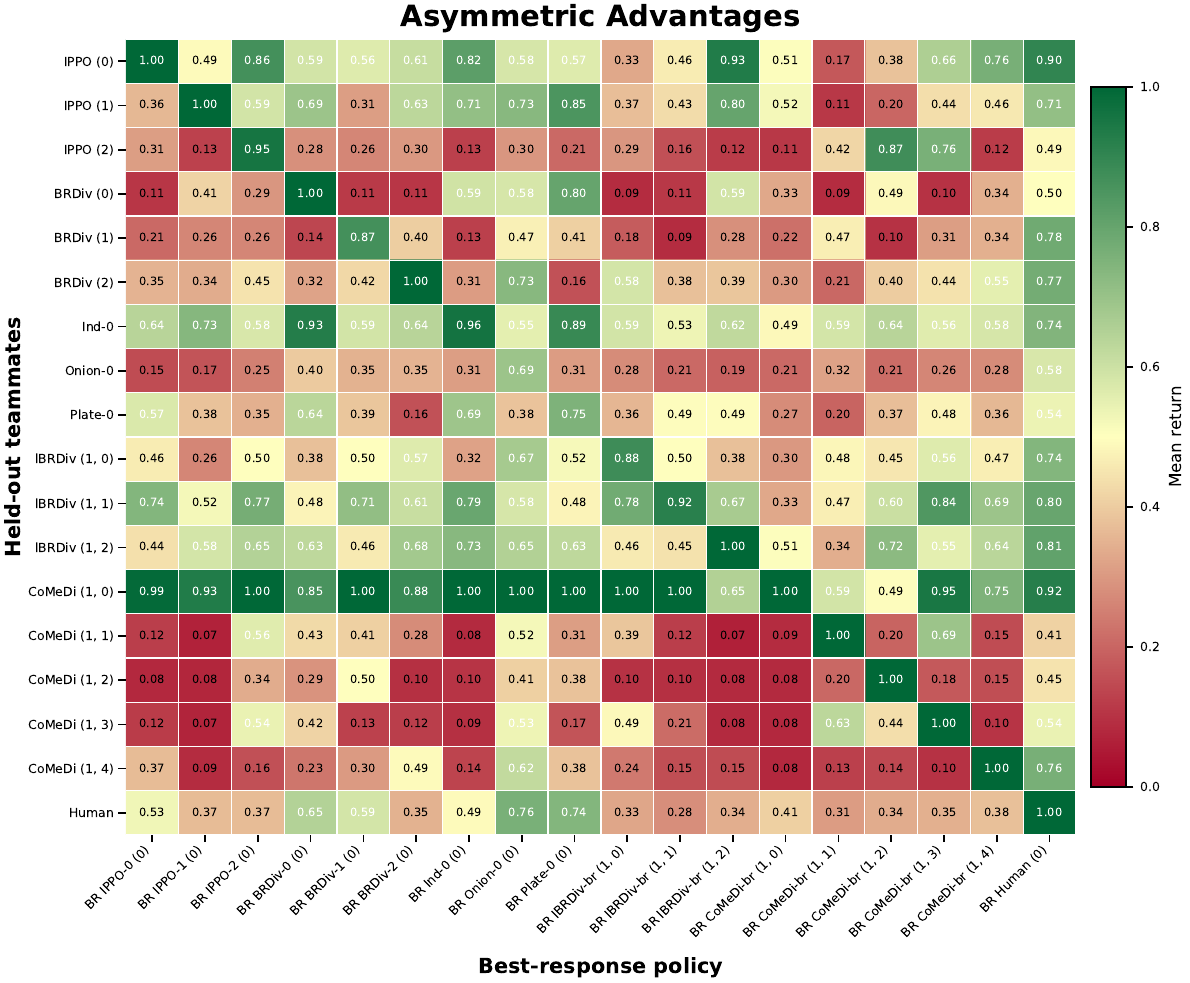}
        \caption{Asymmetric Advantages.}
        \label{fig:overcooked-asymm-xp}
    \end{subfigure}
    \hfill
    \begin{subfigure}[t]{0.48\linewidth}
        \centering
        \includegraphics[width=\linewidth]{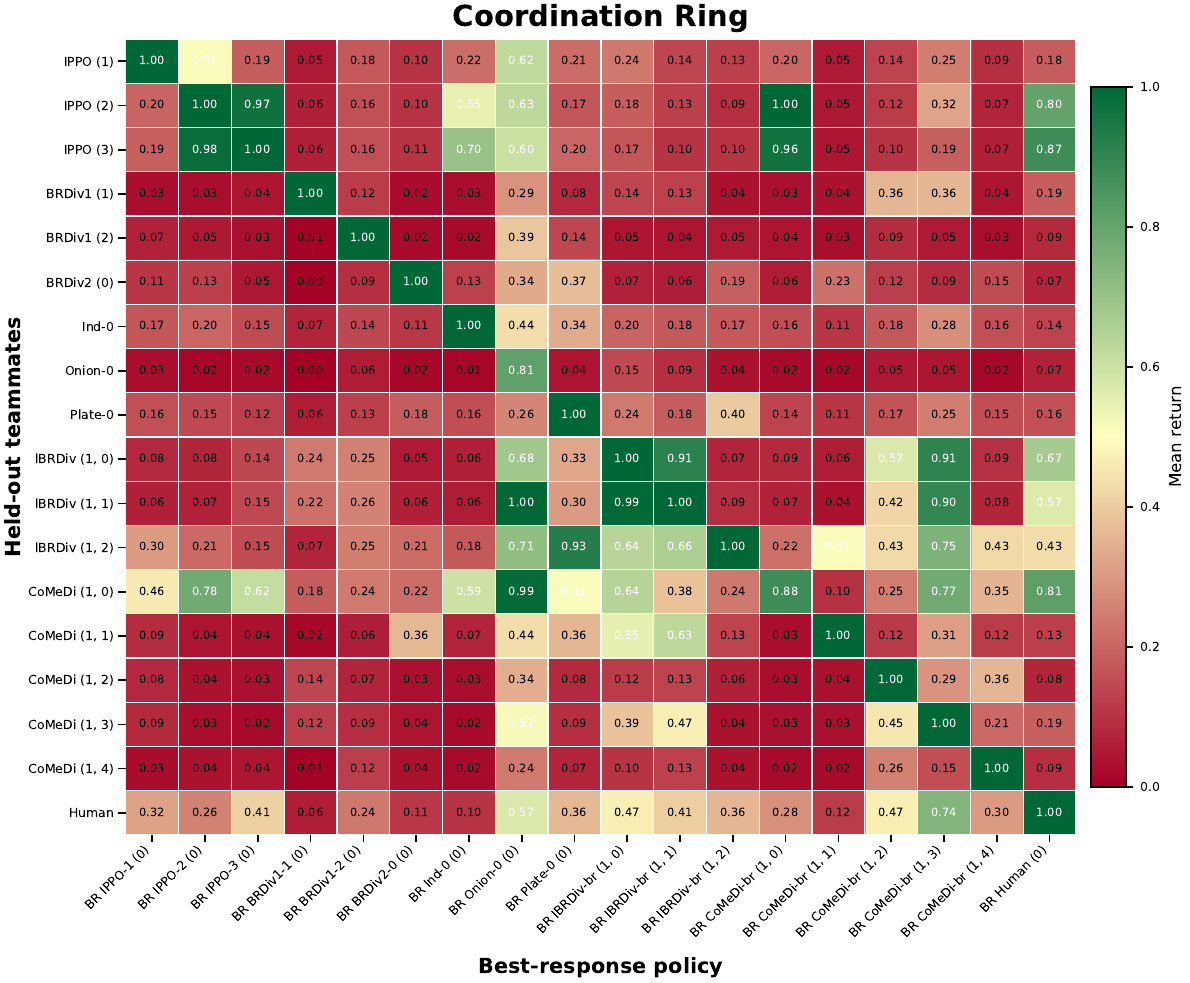}
        \caption{Coordination Ring.}
        \label{fig:overcooked-coord-ring-xp}
    \end{subfigure}
    \vspace{1em}
    \begin{subfigure}[t]{0.48\linewidth}
        \centering
        \includegraphics[width=\linewidth]{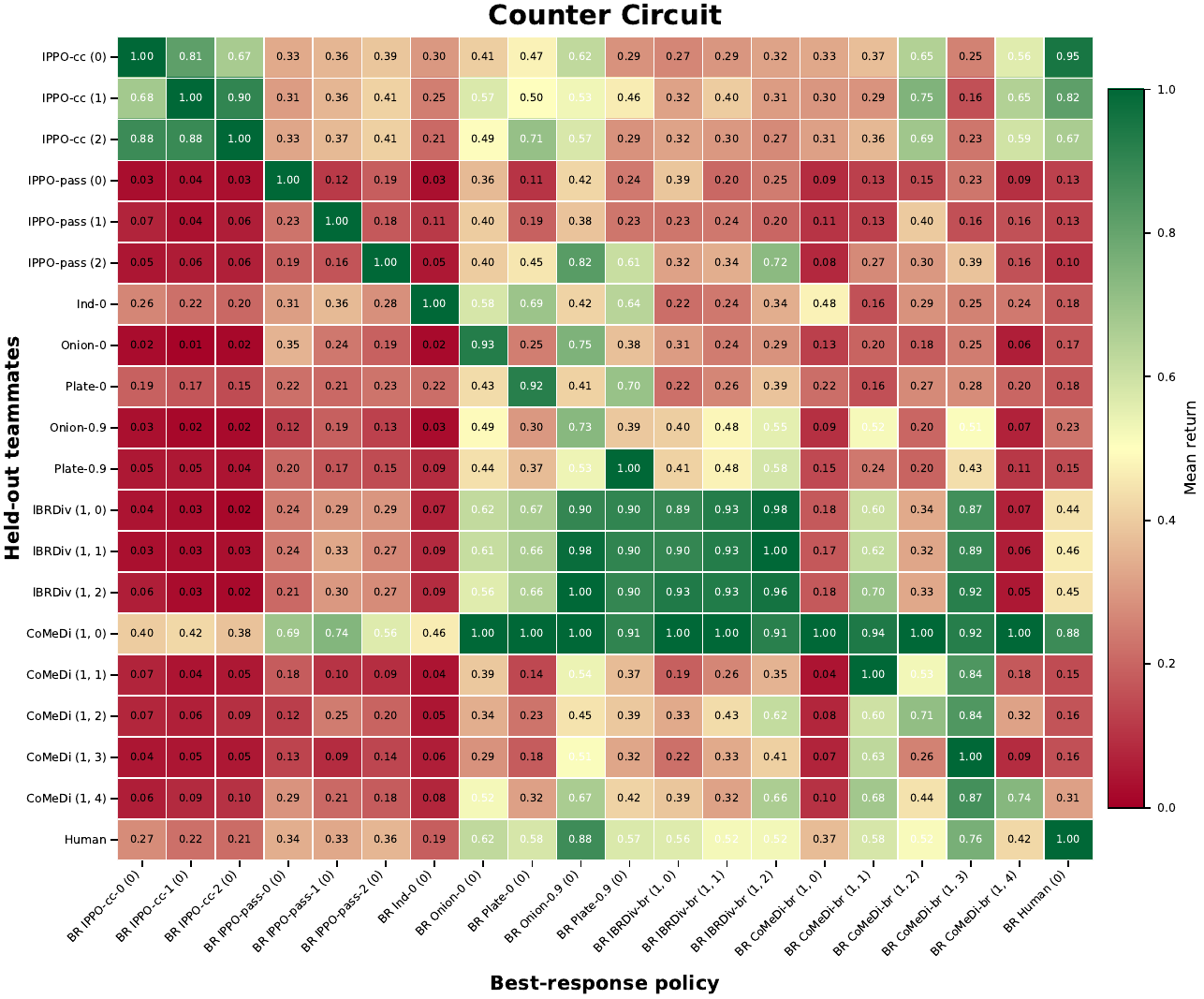}
        \caption{Counter Circuit.}
        \label{fig:overcooked-counter-circuit-xp}
    \end{subfigure}
    \hfill
    \begin{subfigure}[t]{0.48\linewidth}
        \centering
        \includegraphics[width=\linewidth]{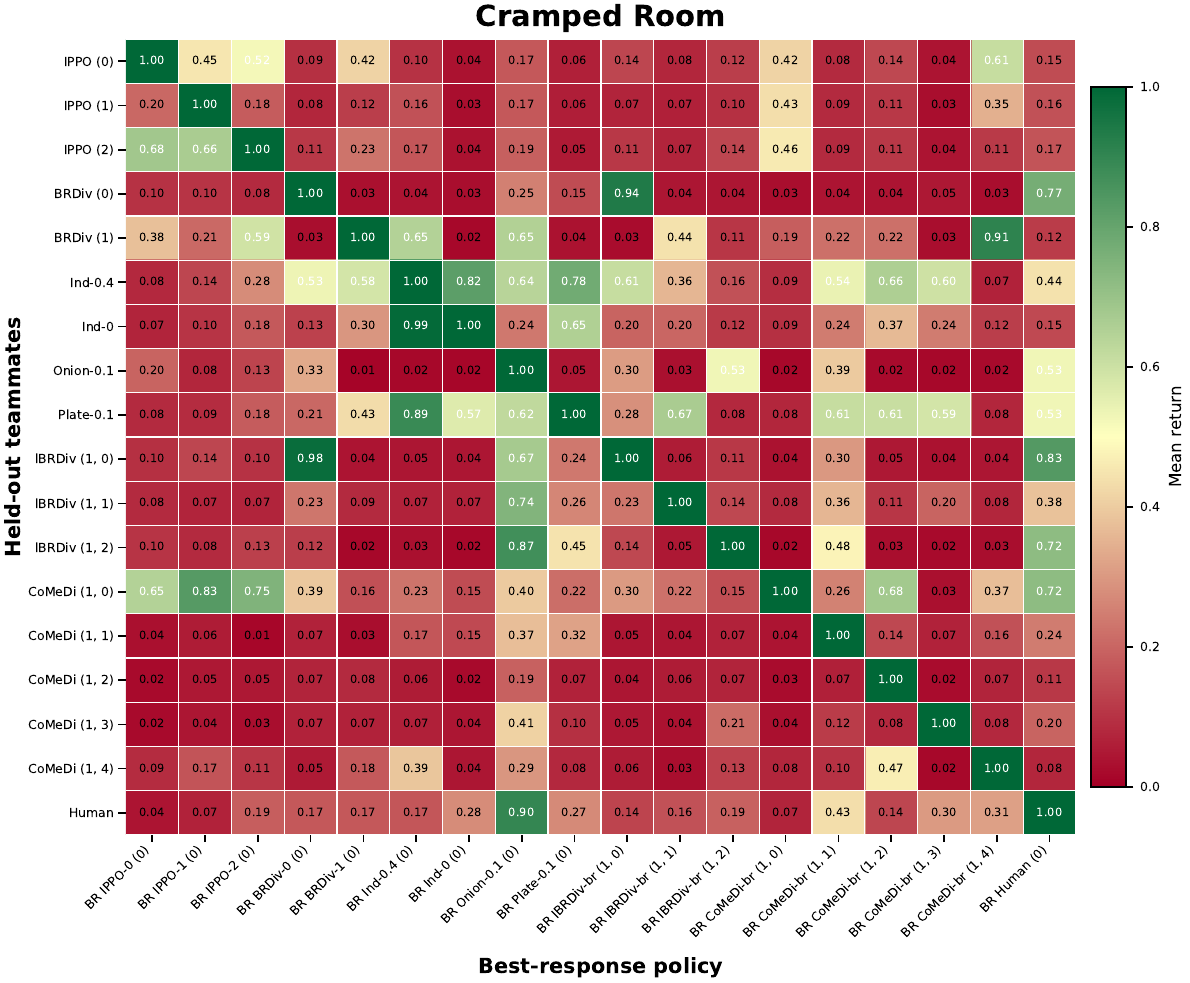}
        \caption{Cramped Room.}
        \label{fig:overcooked-cramped-room-xp}
    \end{subfigure}
    \vspace{1em}
    \begin{subfigure}[t]{0.48\linewidth}
        \centering
        \includegraphics[width=\linewidth]{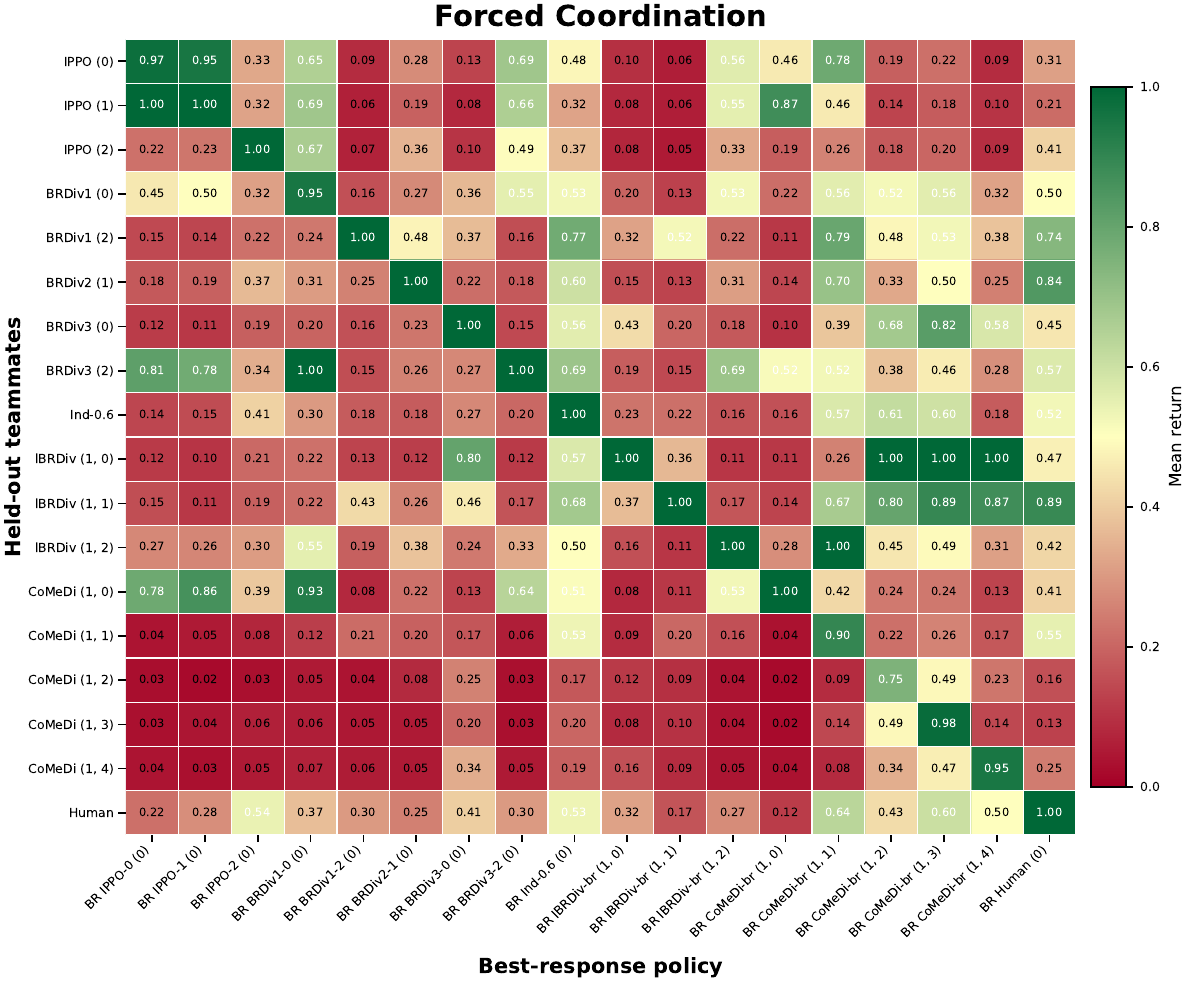}
        \caption{Forced Coordination.}
        \label{fig:overcooked-forced-coord-xp}
    \end{subfigure}
    \caption{\textbf{Overcooked cross-play matrices using normalized return.} Rows are evaluation teammates and columns are best-response policies. Diagonal entries reflect matched best-response performance; off-diagonal structure indicates the degree of transferability across teammate types.}
    \label{fig:overcooked-xp}
\end{figure}

\paragraph{Cross-play matrix construction.}
We evaluate every evaluation teammate against every BR policy, yielding an $N \times N$
XP matrix per task, where rows index evaluation teammates and columns index BR policies. We report normalized episode return based on estimated BR return bounds.
XP matrices for LBF, Overcooked, and Hanabi are shown in Figs.~\ref{fig:lbf-xp-return}, \ref{fig:hanabi_xp}, and \ref{fig:overcooked-xp}.
Diagonal entries are generally stronger than off-diagonal entries, indicating that most evaluation teammates require distinct response policies. 
Some off-diagonal blocks remain, suggesting partial overlap in teammate behavior or transferability across BR policies, especially within agent types (e.g., PPO). 

\paragraph{Upper bound estimation.}
The XP matrix is also used to refine the empirical upper bounds for normalization. For each evaluation teammate, the upper bound is updated if any XP score exceeds the previously recorded value. These bounds remain empirical, as they depend on the particular set of BR policies evaluated.
The resulting upper bound estimates are shown in Fig.~\ref{fig:br_returns_from_config} and correspond to the blue bars in Fig.~\ref{fig:br_return_delta}. 


\section{t-SNE Behavioral Diversity Details}
\label{app:tsne_details}

\begin{figure}[H]
    \centering
    \includegraphics[width=0.55\linewidth]{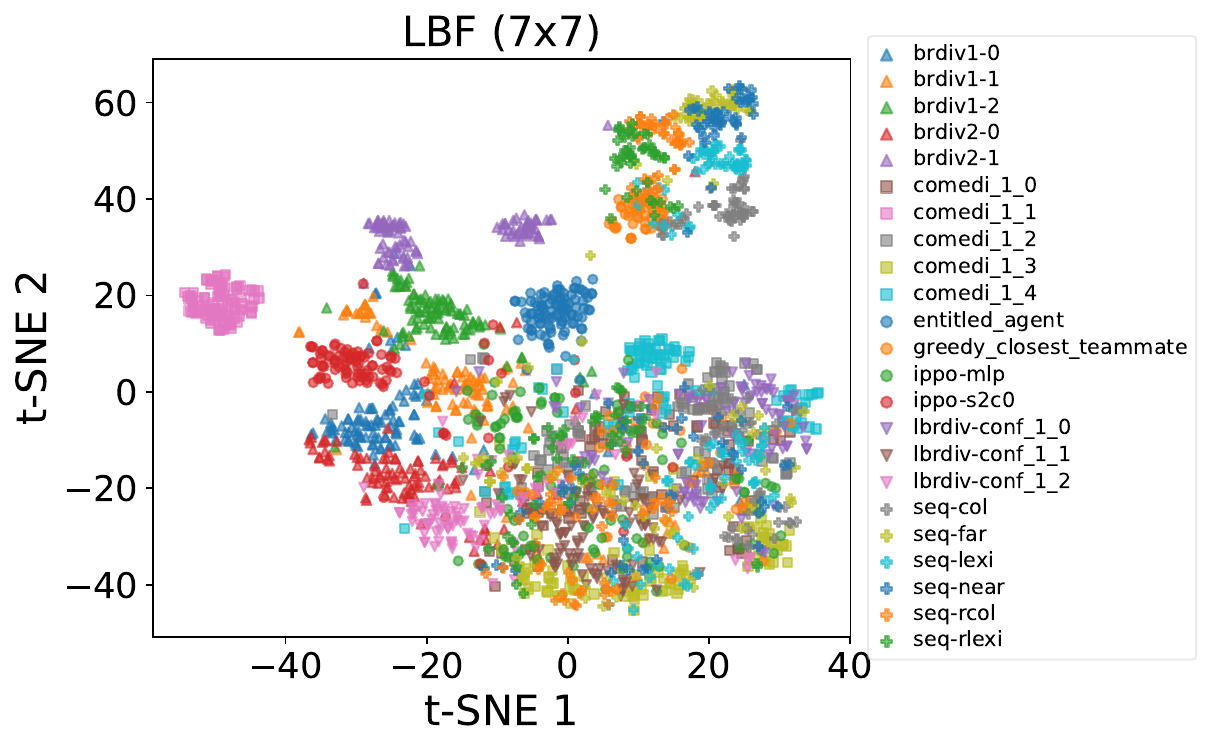}
    \caption{\textbf{Representations of Behavior Trajectories for RL and Heuristic Teammates on LBF 7x7.}}
    \label{fig:lbf-7x7-tsne-visualization}
\end{figure}

To visualize behavioral diversity, we train a classifier on joint interaction data (observations and actions from both the ego agent and the evaluation teammate) to predict the identity of the teammate--best-response pair.
The learned latent representations from this classifier are then projected into two dimensions using t-SNE~\citep{vandermaaten2008tsne}.
Each point in Fig.~\ref{fig:lbf-7x7-tsne-visualization} corresponds to a single episode trajectory, colored by teammate identity; the embeddings are generated by running each evaluation teammate with its own best-response partner.

The classifier is an LSTM encoder that outputs the latent representation of a trajectory, followed by a linear dense layer that maps the latent representation to the predicted teammate--best-response pair.
In order to train the classifier, we collect a fixed number of trajectories from every pair of evaluation teammates and best-responses to evaluation teammates to ensure that the classifier sees a fixed amount of data per pair.
For the training data, we collected 1024 trajectories per pair for LBF 7x7 and 48 trajectories per pair for Coordination Ring.
For the testing data, we collected 100 trajectories per pair for LBF 7x7 and 20 trajectories per pair for Coordination Ring.

\FloatBarrier
\section{Benchmark Study Details}
\label{app:benchmark_details}

We describe the benchmark study in further detail in this section, and present supplemental results. 

For the purposes of the benchmark, tasks are categorized as ``easy'' and ``hard'' based on the general number of timesteps required for MARL algorithms to converge. 
The LBF tasks, Cramped Room, and Mini Hanabi are categorized as easy, while Coordination Ring is categorized as hard.

\subsection{Hyperparameter Sweep}
\label{app:hp_sweep}

We report additional details of the procedure not already stated in the main paper. 
\paragraph{Procedure}
We used Weights \& Biases hyperparameter sweeps.
The network architecture was fixed for all methods. 
The sweeps varied the unique hyperparameters introduced by each algorithm, as well as the learning rate, entropy coefficient, and clip parameter from the base RL algorithm. The remaining hyperparameters from the base RL algorithm were held fixed and set to the same value for all methods.
For each algorithm and task, 140 hyperparameter configurations were swept using random search, with the exception of FCP, which was tuned via grid search over all 90 available configurations, and CoR, where a Bayesian sweep was performed.
The configuration with the highest mean return on an independent validation set consisting of IPPO, LBRDiv, BRDiv, and CoMeDi teammates was selected.

\textit{Ego agent methods} (PPO, LIAM, MeLIBA) were trained for 7M timesteps on easy tasks and 15M timesteps on harder tasks, and were provided FCP partners as teammates.
\textit{Teammate generation} methods were allocated 100M timesteps for easy tasks and 200M timesteps for harder tasks.
Population sizes were varied such that the largest population size exhausted the full budget.
Ego agents were subsequently trained on the fixed generated teammates for 30M steps (easy) and 60M steps (hard), using the same architecture and hyperparameters across methods.
\textit{Unified methods}---which simultaneously train teammates and ego agents---were allocated 130M timesteps on easy tasks and 260M timesteps on harder tasks.

\paragraph{Hyperparameter Return Distributions}

To visualize the effects of hyperparameter variation, we plot the distribution of final normalized returns for each algorithm--task combination.
\begin{figure}[ht]
    \centering
    \includegraphics[width=0.9\linewidth]{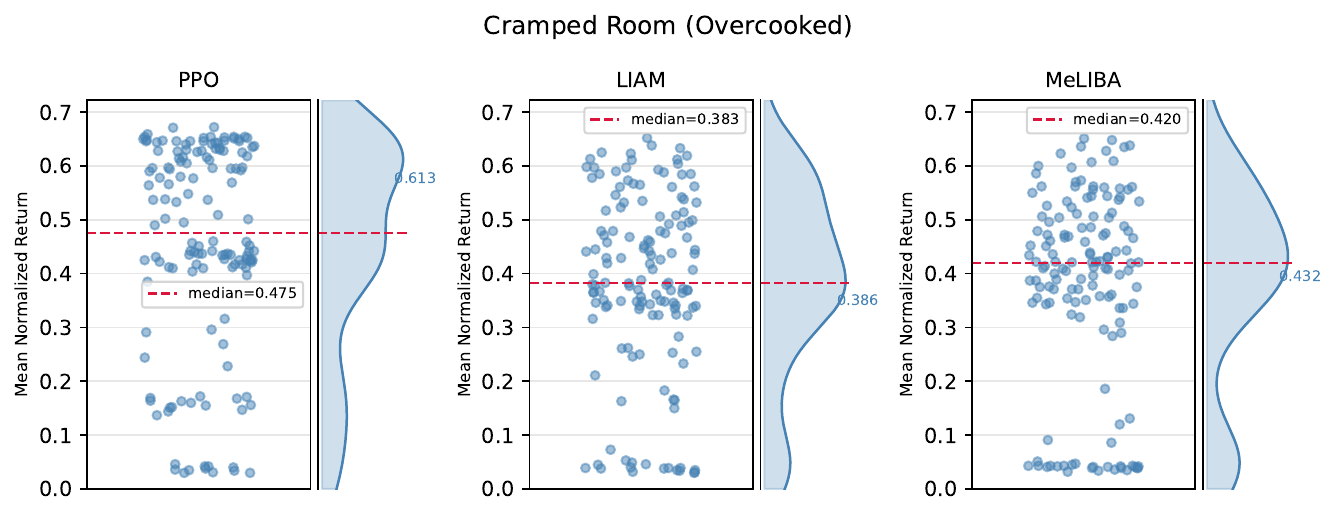}
    \caption{\textbf{Hyperparameter sweep return distributions for ego methods on Cramped Room.}}
    \label{fig:sweep_dist_ego_cramped_room}
\end{figure}

\begin{figure}[ht]
    \centering
    \begin{subfigure}[b]{\linewidth}
        \centering
        \includegraphics[width=\linewidth]{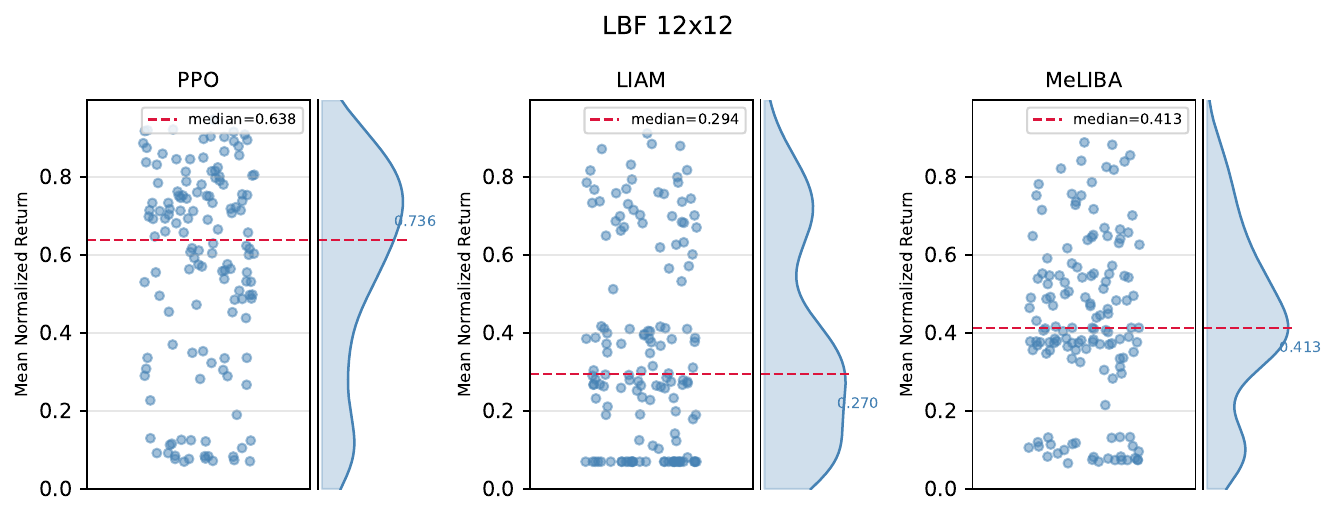}
        \caption{Ego methods.}
    \end{subfigure}
    \\[0.5em]
    \begin{subfigure}[b]{\linewidth}
        \centering
        \includegraphics[width=\linewidth]{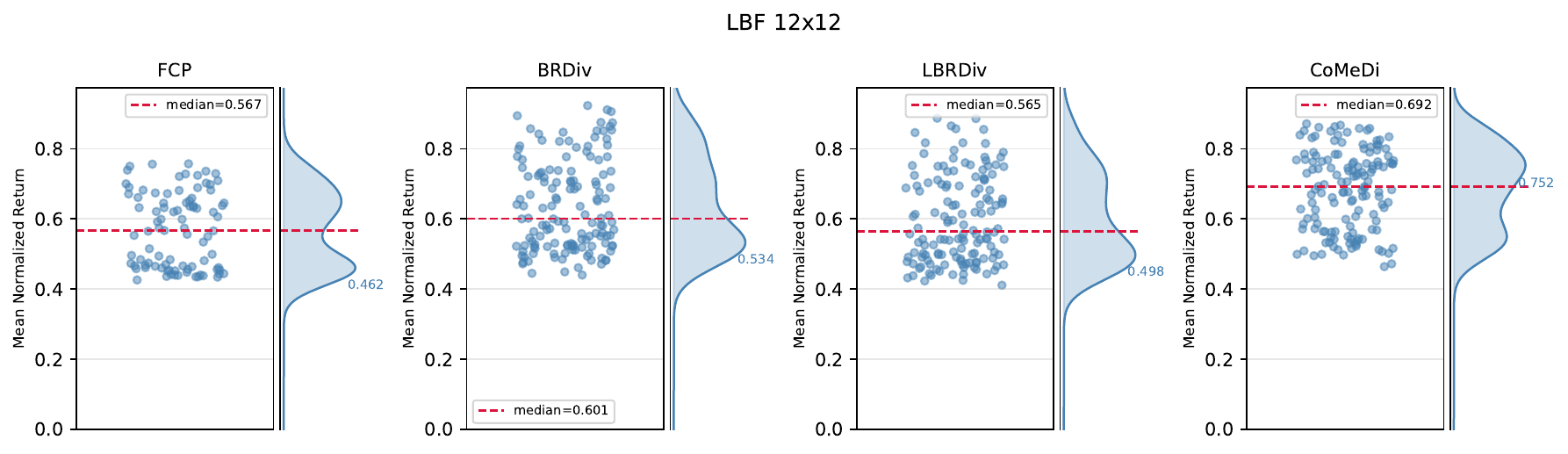}
        \caption{Teammate generation methods.}
    \end{subfigure}
    \\[0.5em]
    \begin{subfigure}[b]{\linewidth}
        \centering
        \includegraphics[width=0.78\linewidth]{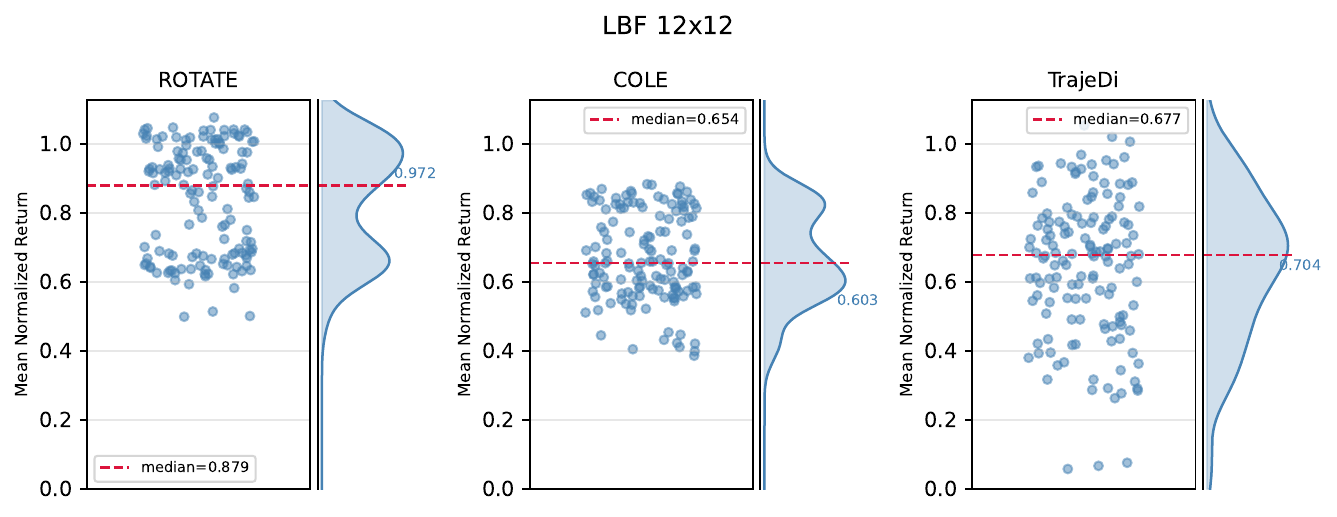}
        \caption{Unified methods.}
    \end{subfigure}
    \caption{\textbf{Hyperparameter sweep return distributions on LBF 12x12.}}
    \label{fig:sweep_dist_lbf}
\end{figure}

\subsection{Benchmark Study}
\label{app:benchmark_procedure}

This section describes the detailed benchmarking procedure, and presents supplemental results. Due to the large number of compared algorithms, tasks, and hyperparameters, we do not directly include them in the paper. We refer the interested reader to the open-sourced codebase's \texttt{benchmark\_expts} branch to view hyperparameter configurations and reproduce experiments. 

\paragraph{Procedure}
The benchmark study uses a 50\% larger timestep budget than the hyperparameter sweep for teammate generation and unified methods, and a 4x larger budget for ego agent training methods, to better allow each method to converge.

Ego agents are trained for 30M timesteps on easy tasks and 60M timesteps on harder tasks, against training populations generated by FCP and CoMeDi. All ego agents share the same recurrent S5 agent architecture~\citep{lu_s5_2023}. 

Teammate generation is allocated 165M timesteps for easy tasks and 330M timesteps for harder tasks, with population sizes varied to exhaust the budget.
Ego agents are then trained on the fixed generated populations for 30M steps (easy) and 60M steps (hard), for a total of 195M and 390M timesteps respectively.
All teammate generation algorithms are compared by training a PPO ego agent on the generated population. 
For fair comparison, the PPO ego agents all use identical architectures and learning configurations. 
Unified methods are allocated 195M total timesteps on easy tasks and 390M on harder tasks.
\subsubsection{Supplemental Figures}

This section provides the supplemental figures previously referenced in Section~\ref{sec:experiments}.

\begin{figure}[ht]
    \centering
    \includegraphics[width=1.0\linewidth]{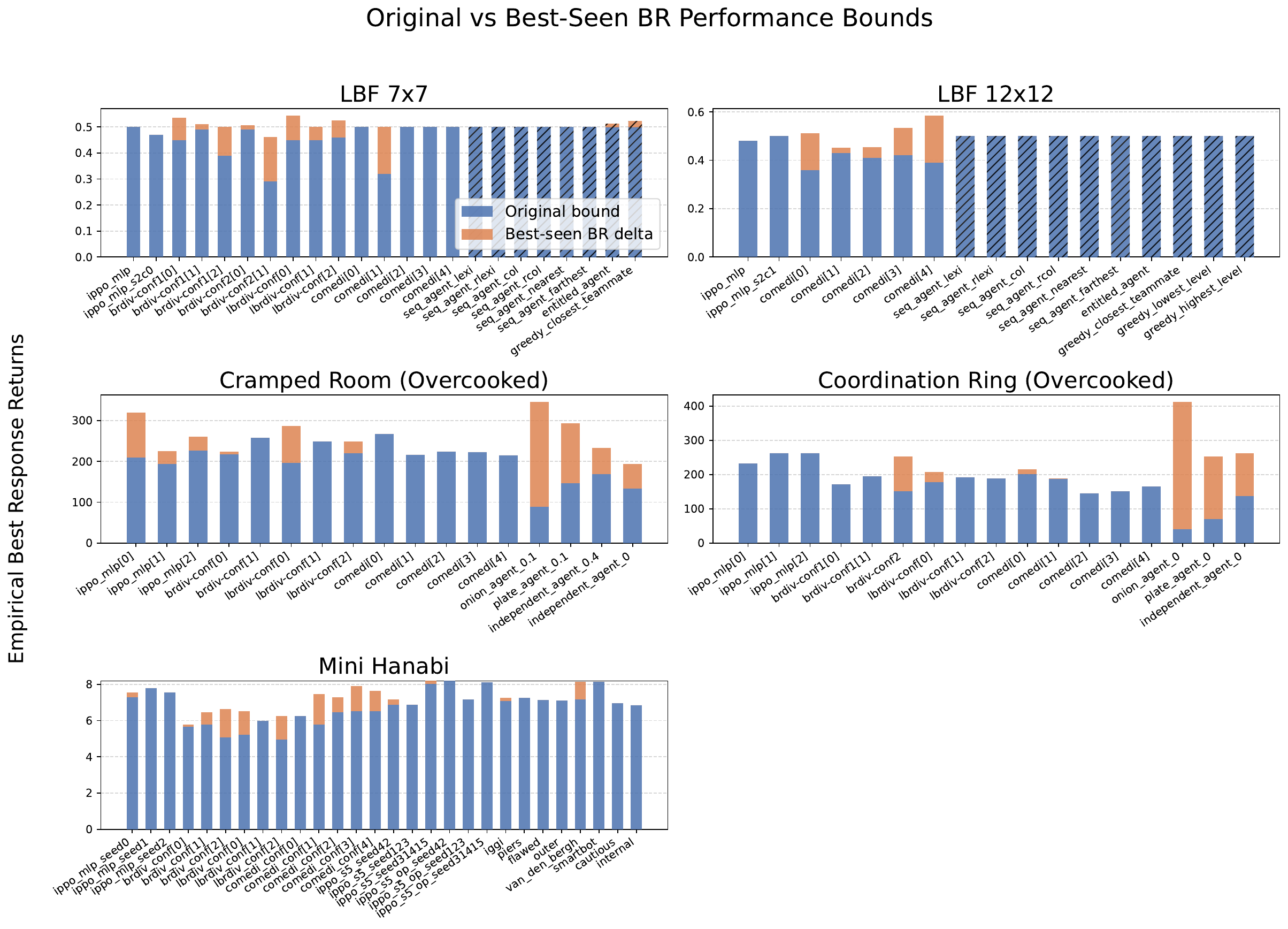}
    \caption{\textbf{BR Return Bounds Gap.} AHT agents can sometimes improve the empirical BR estimates derived by RL.}
    \label{fig:br_return_delta}
\end{figure}

\begin{figure}[ht]
    \centering
    \includegraphics[width=\linewidth]{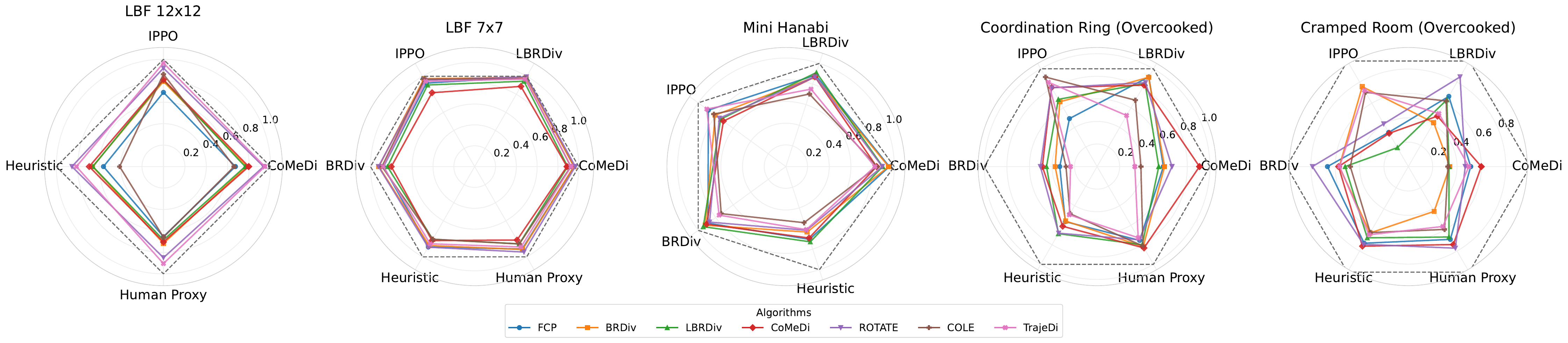}
    \caption{Teammate generation and unified methods performance, broken down by RL agent type.}
    \label{fig:by_agent_type_teamgen_unified}
\end{figure}

\begin{figure}[p]
    \centering
    \includegraphics[width=0.98\linewidth]{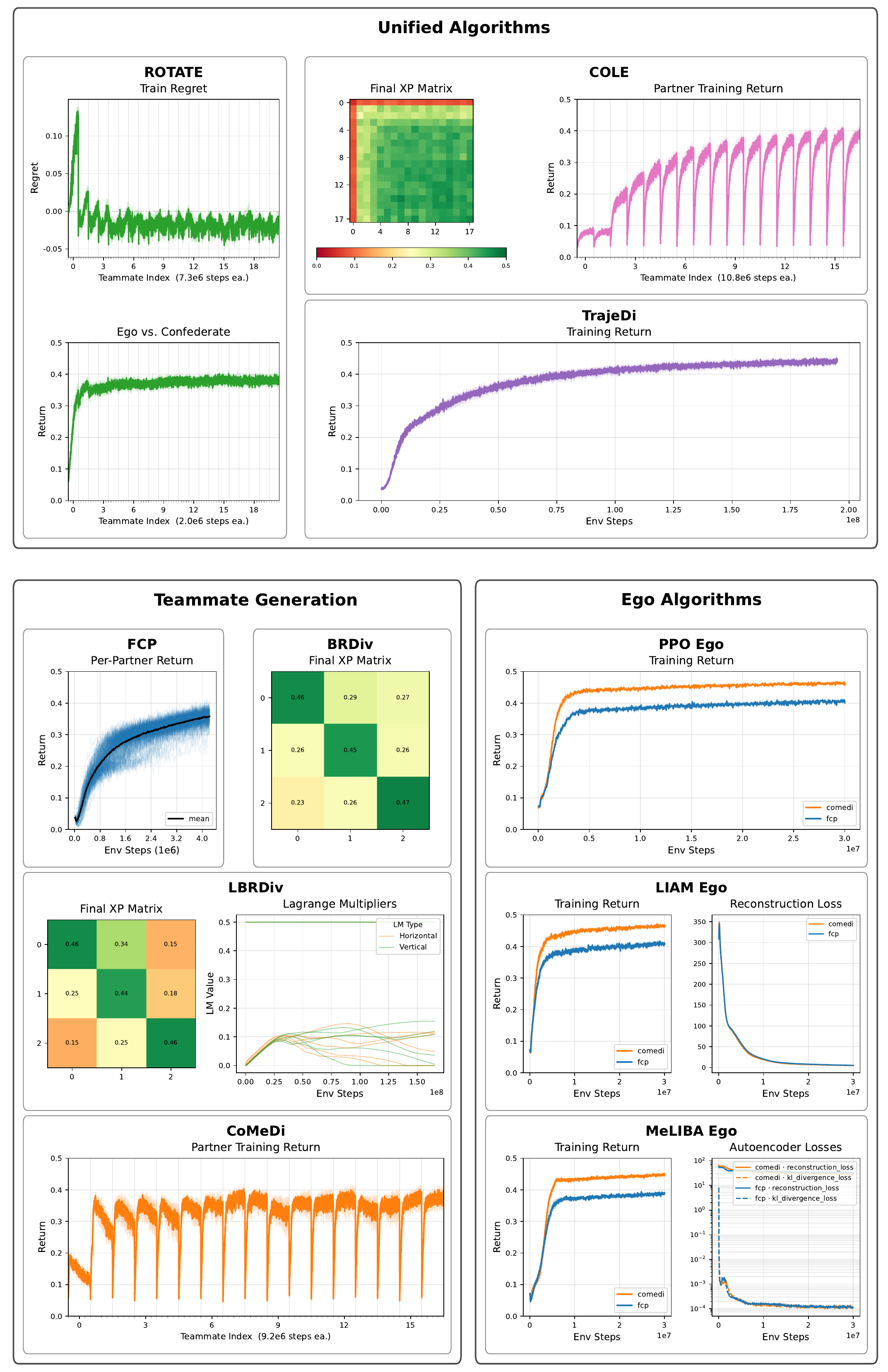}
    \caption{Training statistics for all methods on LBF 12x12.}
    \label{fig:train_loss_curves}
\end{figure}

Fig.~\ref{fig:br_return_delta} displays the improvement in BR return bounds attained by AHT agents, as described in Section~\ref{sec:exp:challenges_br}. 
Fig.~\ref{fig:train_loss_curves} presents key training and loss statistics for all algorithms in the JaxAHT library, on the LBF 12x12 domain.
Fig.~\ref{fig:by_agent_type_teamgen_unified} displays the returns of teammate generation and unified methods, broken down by RL agent type.


\FloatBarrier
\section{Algorithms Overview}
\label{app:alg_overview}

We summarize the learning-based ad hoc teamwork (AHT) algorithms evaluated in our benchmark.
The methods fall into three broad categories.
First, standard ego-agent baselines train a single ego policy against either a fixed teammate distribution or an implicit partner model.
Second, two-stage teammate generation methods first construct a population of training teammates and then train an ego agent against this fixed population.
Third, unified methods simultaneously train teammates and ego agents. 

\paragraph{PPO ego.}
The PPO ego baseline trains an ego policy using Proximal Policy Optimization (PPO)~\citep{Schulman2017ProximalPO}.
In our setting, the ego agent is trained directly against the available training teammate distribution for each task.
This baseline provides a simple reference point for evaluating whether more specialized AHT and teammate-generation methods improve robustness to evaluation partners.

\paragraph{MeLIBA.}
MeLIBA is a meta-learning approach for interactive Bayesian reinforcement learning~\citep{zintgraf2021meliba}.
It is designed to support adaptation by learning representations that allow an agent to infer task- or partner-specific information from interaction history.
In the AHT setting, MeLIBA serves as a learned adaptation baseline that conditions behavior on observed experience with the current teammate.

\paragraph{LIAM.}
LIAM learns latent representations for agent modeling under partial observability~\citep{papoudakis2020liam}.
The method uses interaction trajectories to infer hidden information about other agents and conditions the ego policy on this inferred representation.
In our benchmark, LIAM represents a partner-modeling baseline for adapting to unfamiliar teammates through learned latent inference.

\paragraph{Fictitious Co-Play (FCP).}
Fictitious Co-Play (FCP) is a two-stage AHT method designed to improve robustness to independently trained partners~\citep{strouse2021fcp}.
In the first stage, a population of teammate policies is generated by running independent cooperative MARL training, typically with different random seeds and checkpoints.
In the second stage, the ego agent is trained against this fixed teammate population.
FCP can generalize well to teammates that resemble independently trained agents, but its coverage is limited by the behavioral diversity of the generated teammate pool.

\paragraph{BRDiv.}
Best-Response Diversity (BRDiv) is a two-stage teammate-generation method that constructs a population of confederate teammate policies together with corresponding best-response policies~\citep{rahman2023brdiv}.
BRDiv maintains a cross-play matrix between generated teammates and their best responses.
The objective encourages high matched teammate--best-response performance and low mismatched cross-play performance.
This encourages the generated teammates to require distinct best-response strategies.
After the teammate population is generated, an ego agent is trained against the fixed confederate population.

\paragraph{LBRDiv.}
LBRDiv extends the BRDiv framework by seeking a compact teammate set that provides sufficient behavioral coverage for robust AHT training~\citep{rahman2024lbrdiv}.
Rather than simply increasing the number of generated teammates, LBRDiv focuses on identifying a minimum coverage set of teammates.
This makes it a useful comparison for evaluating whether a smaller but more strategically selected teammate population can provide similar or better generalization than larger generated populations.

\paragraph{CoMeDi.}
CoMeDi is a two-stage teammate-generation method designed to generate diverse conventions while avoiding the self-sabotage that can arise from purely adversarial diversity objectives~\citep{sarkar2023comedi}.
It trains teammate policies sequentially.
For each new teammate, the objective encourages strong self-play performance, low compatibility with previously generated teammates, and high mixed-play performance.
The mixed-play term exposes the new teammate to states that arise when conventions are partially mismatched, encouraging it to remain cooperative outside its own self-play trajectory distribution.
After the teammate population is generated, an ego agent is trained against the fixed population.

\paragraph{ROTATE.}
ROTATE is an open-ended AHT algorithm that alternates between teammate generation and ego-agent training~\citep{wang2025rotateregretdrivenopenendedtraining}.
At each iteration, ROTATE generates a new teammate that maximizes the current ego agent's cooperative regret, defined as the gap between the return achievable by a best response to that teammate and the return achieved by the current ego agent.
The ego agent is then trained against the accumulated buffer of generated teammates.
Unlike two-stage teammate-generation methods, ROTATE generates new teammates in response to the ego agent's current weaknesses rather than constructing a fixed population before ego training begins.
ROTATE also optimizes a per-state regret objective, which is intended to reduce the sabotage-inducing properties of regret-based teammate generation objectives.

\paragraph{TrajeDi.}
Trajectory Diversity (TrajeDi) is a population-based diversity method originally proposed for zero-shot coordination~\citep{pmlr-v139-lupu21a}.
TrajeDi trains a population of policies while regularizing them to induce diverse trajectory distributions.
Because the diversity objective is defined at the trajectory level, it can capture behavioral differences that may not be visible from action distributions alone.
In our benchmark, TrajeDi serves as an information-theoretic unified baseline that emphasizes broad behavioral coverage rather than explicit best-response diversity.

\paragraph{COLE.}
Cooperative Open-Ended Learning (COLE) is a unified framework for zero-shot coordination that addresses cooperative incompatibility within a population of strategies~\citep{li2023cole}.
Rather than only maximizing behavioral diversity, COLE evaluates cooperative relationships among policies using graph-theoretic notions of compatibility.
At each generation, the method identifies weaknesses in the current population and adds a new strategy that improves the population's ability to coordinate with different partners.
In this sense, COLE is related to open-ended teammate generation, but its central focus is maintaining cooperative compatibility across the policy population rather than only increasing diversity.

\section{Evaluation Teammate Descriptions}
\label{app:teammate_descriptions}

A key contribution of JaxAHT is a comprehensive suite of heuristic teammates spanning all three environments. 
The Hanabi heuristic suite consists of eight rule-based agents drawn from two sources: seven agents from the~\citet{waltonrivers2017} rule-based suite, and SmartBot, the strongest teammate in the suite~\citep{smartbot2019}. To our knowledge, JaxAHT provides the first JAX implementation of these rule-based agents.  
The LBF heuristic teammate suite consists of three agent types spanning 12 strategies. The Overcooked-v1 heuristic teammates comprise three role-based agents, including role specialists such as Onion and Plate. 

Table~\ref{tab:heuristic_suite} summarizes the human-designed heuristic teammates used across the benchmark environments.

\begin{table}[ht]
\centering
\small
\setlength{\tabcolsep}{4pt}
\renewcommand{\arraystretch}{1.12}
\begin{tabular}{@{}ll>{\raggedright\arraybackslash}p{7.2cm}@{}}
\toprule
Env & Agent & Behavior \\
\midrule
\textbf{Hanabi}
& Cautious & Derived from human gameplay; avoids plays that can lose lives. \\
& IGGI & Extends Cautious with conservative play and predictable discards. \\
& Piers & Risk-taking extension of IGGI, including Hail Mary plays when the deck is empty. \\
& Internal~\citep{osawa2015hanabi} & Tracks hints received about its own hand, without maintaining the full history of hints given to its partner. \\
& Outer~\citep{osawa2015hanabi} & Tracks hints about its own hand and hints given to its partner, using the latter to avoid redundant tells. \\
& Van Den Bergh & Rule-based agent whose rule ordering was found by a genetic algorithm~\citep{vdb2017}. \\
& Flawed & Uses generally informed play but lacks non-random tell rules and includes a risky play rule. \\
& SmartBot~\citep{smartbot2019} & Convention-based agent that tracks per-card knowledge from accumulated hint history, and prioritizes safe plays. The strongest heuristic teammate in the suite. \\
\midrule
\textbf{LBF}
& Sequential & Plans a food order before the episode using \texttt{lexicographic}, \texttt{reverse\_lexicographic}, \texttt{column\_major}, \texttt{reverse\_column\_major}, \texttt{nearest\_agent}, or \texttt{farthest\_agent}. \\
& Greedy & Selects the next uneaten food online using \texttt{closest\_self}, \texttt{closest\_teammate}, \texttt{closest\_avg}, \texttt{lowest\_level}, or \texttt{highest\_level}. \\
& Entitled & Chooses a pair of adjacent squares for each food and waits until the teammate occupies one of them before approaching the other square and loading. \\
\midrule
\textbf{Overcooked-v1}
& Onion & Collects onions and places them in pots, with a configurable probability of staging onions on counters. \\
& Plate & Gets plates, plates ready soup, and delivers dishes, with a configurable probability of staging plates on counters. \\
& Independent & Executes the full onion--plate--deliver pipeline, with configurable probabilities of staging onions or plates on counters. \\
\bottomrule
\end{tabular}
\caption{\small\textbf{Heuristic teammate suite.} The hand-designed teammates span convention-based Hanabi policies, LBF food-targeting strategies, and Overcooked-v1 role-specialist policies; the listed behaviors define the coordination demands used in held-out evaluation.}
\label{tab:heuristic_suite}
\end{table}

\section{LBF Human Data Collection}
\label{app:human_data_lbf}

To create human-data based agents for LBF, we collect and provide a dataset of human gameplay against various agents.
The data collection study was approved by \ifpreprint The University of Texas at Austin Institutional Review Board (IRB \#STUDY00008549)\else the institution's IRB\fi. Participants were informed that participation was voluntary and were provided contact information for questions about the study, consistent with IRB requirements.

\subsection{Procedure}
Participants first encounter an instructional interface that explains the rules and mechanics of the online LBF game. These instructions are presented through a popup prior to gameplay to ensure that all participants understand the objectives, controls, and cooperative aspects of the environment.
Each participant then plays a sequence of ten games in a two-player setting consisting of one human participant and an automated LBF bot uniformly sampled from a set of bots. Every game lasts for a maximum of 50 steps, where a step corresponds to a single action input from the human player. To assist participants in pacing their gameplay, the interface displays the number of steps completed out of the total 50 allowed in the current game.

The first two games serve as warm-up rounds intended to familiarize participants with the gameplay dynamics and interface. These warm-up games are selected randomly from the set of possible game configurations, and data from these rounds are not included in the final dataset. Their sole purpose is to allow participants to gain experience with the mechanics of Level-Based Foraging before the experimental data collection begins. Following the warm-up phase, participants complete eight additional games from which data are collected. 

\subsection{Variants}
The experiment includes four distinct game configurations, the cross product of grid sizes (7x7 and 12x12) and whether or not fruits have different levels. 
All variants had teammates comprising all applicable heuristic agents (sequential, greedy, entitled) and one agent trained via IPPO \citep{dewitt2020independentlearning}. 

\subsection{Prolific}
This study was conducted on Prolific with 180 different participants, each providing 8 episodes, for a total of 1440 episodes across all variants. Participants were compensated via the Prolific platform between 8 and 9 USD per hour.

We filtered out participants whom we deemed not to be playing in good faith if any of the following criteria held: (1) $>$35\% of actions were unsuccessful, (2) $>$15\% of actions were idle (wait), (3) they scored zero in 5 or more of their 8 games.
This left us with 1299 episodes in our dataset. 
The user study dashboard is released with our codebase. 

\begin{figure}[H]
    \centering
    \begin{subfigure}[t]{0.49\linewidth}
        \centering
        \includegraphics[width=\linewidth]{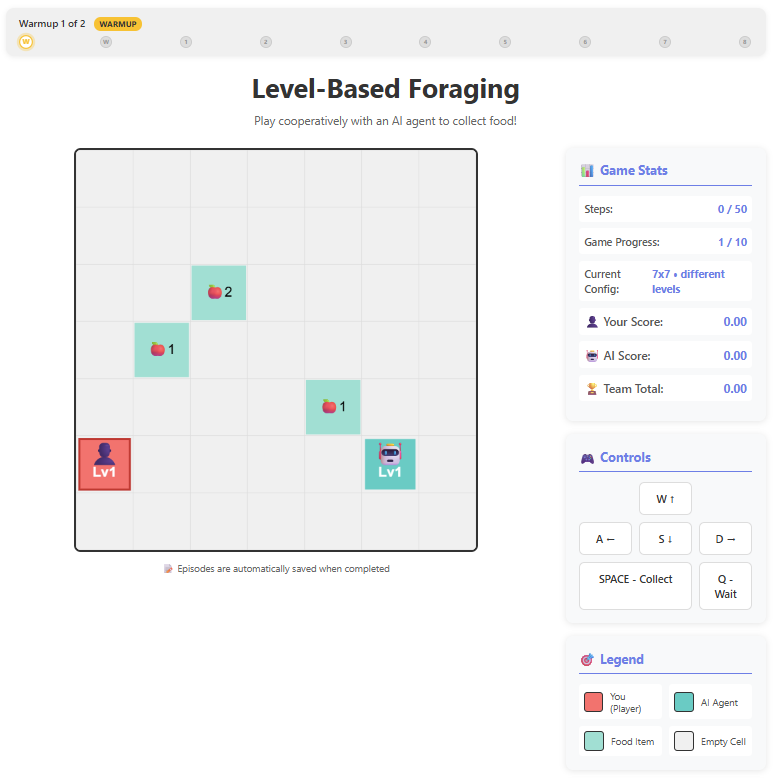}
        \caption{Prolific Study Gameplay.}
        \label{fig:prolific_gameplay}
    \end{subfigure}
    \hfill
    \begin{subfigure}[t]{0.49\linewidth}
        \centering
        \includegraphics[width=\linewidth]{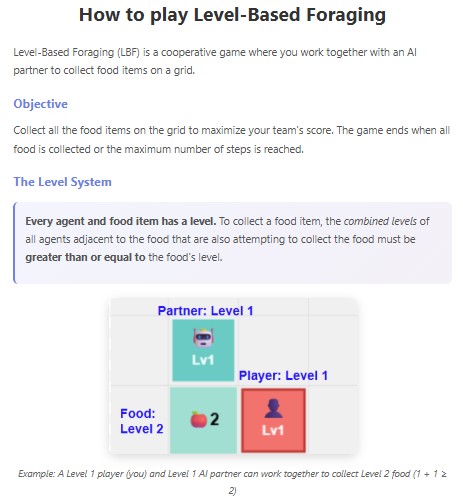}
        \caption{Prolific Study Instructions.}
        \label{fig:prolific_instructions}
    \end{subfigure}
    \caption{Prolific study interface.}
    \label{fig:prolific_study}
\end{figure}

\FloatBarrier
\section{Attributions}
\label{app:attributions}

We gratefully acknowledge the following open-source software, datasets, and model weights used in this work.

\begin{itemize}
    \item \textbf{JaxMARL} \citep{rutherford2024jaxmarl}: Our IPPO implementations build on the JaxMARL codebase. We also use the Overcooked and Hanabi environments provided by this library. (Apache 2.0 License.)

    \item \textbf{Jumanji} \citep{bonnet2023jumanji}: We use the Level-Based Foraging (LBF) environment from this library. (Apache 2.0 License.)

    \item \textbf{AH2AC2} \citep{dizdarevic2025ah2ac2}: We use the publicly released human gameplay data. (Apache 2.0 License.)

    \item \textbf{Overcooked human data} \citep{carroll_utility_2019}: We use the human gameplay data open-sourced alongside the original Overcooked Benchmark. (MIT License.)

    \item \textbf{Off-Belief Learning (OBL) policy weights} \citep{hu2021offbelieflearning}: We use the pretrained OBL policy weights released under a CC BY 4.0 License. We additionally use a third-party Flax port providing OBL at all five belief levels, available at \url{https://huggingface.co/mttga/obl-r2d2-flax}.
\end{itemize}

\section{Compute Infrastructure}
\label{app:compute}

Experiments were performed on various servers, with the configurations listed below.

\begin{itemize}[leftmargin=*]
    \item Intel(R) Xeon(R) Gold 6342 CPUs; NVIDIA A40 GPUs (48GB)
    \item Intel(R) Xeon(R) Gold 6342 CPUs; NVIDIA A100 GPUs (80GB)
    \item Intel(R) Xeon(R) Platinum 8468; NVIDIA H100 GPUs (96GB)
    \item Arm Neoverse-V2 CPU; NVIDIA GH200 (96GB)
    \item AMD EPYC 7763 CPU (Milan); NVIDIA A100 GPUs (40GB)
    \item AMD Ryzen Threadripper PRO 9955WX; NVIDIA RTX 6000 Ada Generation (48 GB)
    \item AMD EPYC; NVIDIA RTX Pro 6000 (Blackwell; 96GB)
\end{itemize}

\end{document}